\documentclass[journal]{IEEEtran}
\usepackage{amsmath,amsfonts}
\usepackage{algorithmic}
\usepackage{algorithm}
\usepackage{array}
\usepackage[caption=false,font=normalsize,labelfont=sf,textfont=sf]{subfig}
\usepackage{textcomp}
\usepackage{stfloats}
\usepackage{url}
\usepackage{verbatim}
\usepackage{graphicx}
\usepackage{booktabs}
\usepackage{multirow}
\usepackage{makecell}
\usepackage[table,xcdraw]{xcolor}

\usepackage[normalem]{ulem}
\useunder{\uline}{\ul}{}
\usepackage{float}
\usepackage[numbers,sort&compress]{natbib}
\usepackage{hyperref}
\begin{document}

% \title{Predicting Globally and Foreseeing Locally: Target-Aware Aerial Video Prediction}
\title{TAFormer: A Unified Target-Aware Transformer for Video and Motion Joint Prediction in Aerial Scenes}

\author{IEEE Publication Technology,~\IEEEmembership{Staff,~IEEE,}
        % <-this % stops a space
\thanks{This paper was produced by the IEEE Publication Technology Group. They are in Piscataway, NJ.}% <-this % stops a space
\thanks{Manuscript received April 19, 2021; revised August 16, 2021.}}

 \author{Liangyu Xu, Wanxuan Lu, Hongfeng Yu, Yongqiang Mao, Hanbo Bi, Chenglong Liu,\\ Xian Sun, \IEEEmembership{Senior~Member,~IEEE,} Kun~Fu, \IEEEmembership{Senior~Member,~IEEE}% <-this % stops a space
\thanks{\textit{(Corresponding authors: Wanxuan~Lu, Hongfeng~Yu)}}

\thanks{Liangyu Xu, Yongqiang Mao, Hanbo Bi, Chenglong Liu, Xian Sun, Kun Fu are with the Aerospace Information Research Institute, Chinese Academy of Sciences, Beijing 100094, China, also with the Key Laboratory of Network Information System Technology (NIST), Aerospace Information Research Institute, Chinese Academy of Sciences, Beijing 100190, China, also with the University of Chinese Academy of Sciences, Beijing 100049, China, and also with the School of Electronic, Electrical and Communication Engineering, University of Chinese Academy of Sciences, Beijing 100049, China (e-mail: xuliangyu21@mails.ucas.ac.cn; maoyongqiang19@mails.ucas.ac.cn; bihanbo21@mails.ucas.edu.cn; liuchenglong20@mails.ucas.ac.cn; sunxian@aircas.ac.cn; kunfuiecas@gmail.com).}
\thanks{Wanxuan Lu and Hongfeng Yu are with the Aerospace Information Research Institute, Chinese Academy of Sciences, Beijing 100094, China, and also with the Key Laboratory of Network Information System Technology (NIST), Aerospace Information Research Institute, Chinese Academy of Sciences, Beijing 100190, China (e-mail: luwx@aircas.ac.cn; yuhf@aircas.ac.cn)}

}

% The paper headers
% \markboth{Journal of \LaTeX\ Class Files,~Vol.~14, No.~8, August~2021}%
% {Shell \MakeLowercase{\textit{et al.}}: A Sample Article Using IEEEtran.cls for IEEE Journals}

% \IEEEpubid{0000--0000/00\$00.00~\copyright~2021 IEEE}
% Remember, if you use this you must call \IEEEpubidadjcol in the second
% column for its text to clear the IEEEpubid mark.

\maketitle

% Existing video prediction methods focus solely on predicting future scenes (video frames) and don't explicitly consider motion states of the target, which is crucial for aerial video interpertation. 

\begin{abstract}
As drone technology advances, using unmanned aerial vehicles for aerial surveys has become the dominant trend in modern low-altitude remote sensing. The surge in aerial video data necessitates accurate prediction for future scenarios and motion states of the interested target, particularly in applications like traffic management and disaster response. Existing video prediction methods focus solely on predicting future scenes (video frames), suffering from the neglect of explicitly modeling target's motion states, which is crucial for aerial video interpretation. To address this issue, we introduce a novel task called Target-Aware Aerial Video Prediction, aiming to simultaneously predict future scenes and motion states of the target. Further, we design a model specifically for this task, named TAFormer, which provides a unified modeling approach for both video and target motion states. Specifically, we introduce Spatiotemporal Attention (STA), which decouples the learning of video dynamics into  spatial static attention and temporal dynamic attention, effectively modeling the scene appearance and motion. Additionally, we design an Information Sharing Mechanism (ISM), which elegantly unifies the modeling of video and target motion by facilitating information interaction through two sets of messenger tokens. Moreover, to alleviate the difficulty of distinguishing targets in blurry predictions, we introduce Target-Sensitive Gaussian Loss (TSGL), enhancing the model's sensitivity to both target's position and content. Extensive experiments on UAV123VP and VisDroneVP (derived from single-object tracking datasets) demonstrate the exceptional performance of TAFormer in target-aware video prediction, showcasing its adaptability to the additional requirements of aerial video interpretation for target awareness.
\end{abstract}


\section{Introduction} \label{sec:introdution}
\IEEEPARstart{W}{ith} the swift advancement in drone technology, the scale of aerial video data is experiencing rapid growth, providing rich information for the analysis of various scenes. Relying solely on perception and tracking\cite{jiao2023transformer,zhang2021moving,cui2021remote} is increasingly inadequate to meet the demand for higher-level intelligent analysis of aerial videos. Particularly in applications such as traffic management and disaster response, foreseeing future scenarios is crucial for informed decision-making and there is an urgent need for accurate prediction of environmental evolution and motion states of the target in aerial scenes.

% Existing relevant research\cite{shi2015convolutional,shi2017deep,wang2018predrnn++,wang2022predrnn,gao2022simvp,tan2022simvp,tan2023temporal} primarily focus on the video prediction task in Computer Vision (CV), which typically involve modeling the spatiotemporal dynamics of historical consecutive frames to predict the global scene in the future (i.e., regressing the pixel values for future frames), as shown in Fig. 1(a). Nevertheless, in many aerial applications, we not only need to predict the global scene but also pay closer attention to the motion states of the interested target. Previous video prediction methods cannot explicitly capture the dynamic behavior of the target, especially when it undergoes complex pose changes in the video sequence. Additionally, video prediction is susceptible to the influence of environmental noise, reducing the predictive accuracy for target details. Some studies\cite{altche2017lstm,wiest2012probabilistic,nikhil2018convolutional,houenou2013vehicle,messaoud2020attention,DBLP:journals/corr/abs-2003-08111,mo2022multi} focus on trajectory prediction for the target of interest, the main limitation of which is a lack of foresight for changes in the global scene. Predicting the motion states of specific targets is often constrained by historical trajectory information and static overhead views of the environment, lacking adaptability to environmental changes.

\begin{figure*}[t]
	\setlength{\abovecaptionskip}{1pt}
	\centering
	\includegraphics[width=0.9\linewidth]{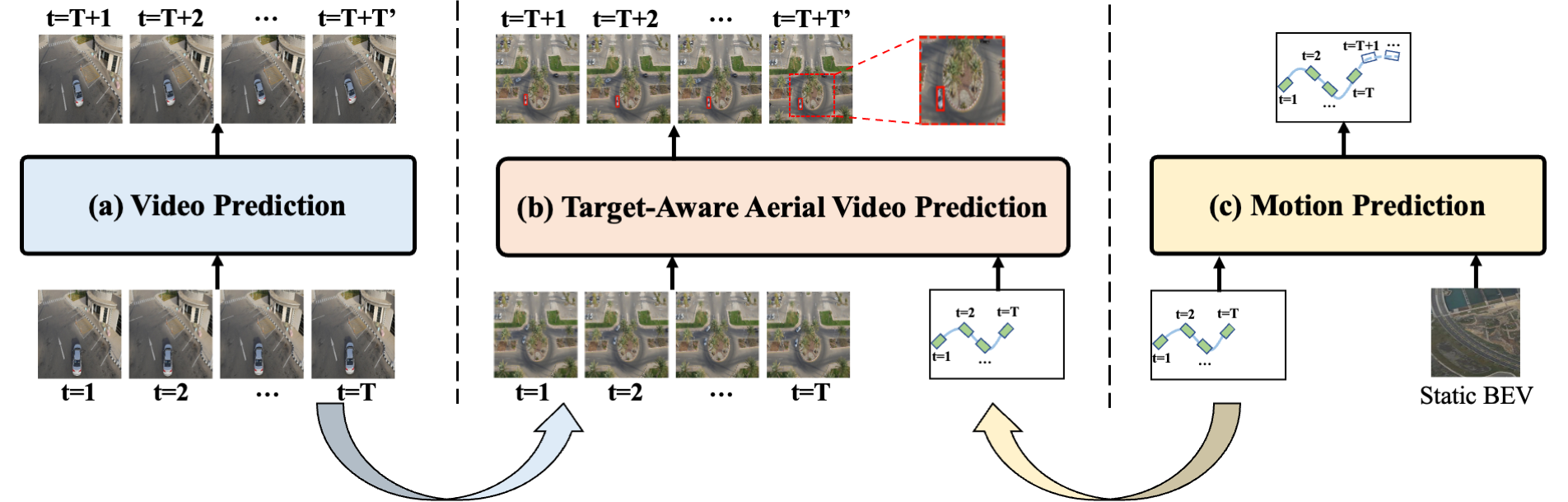}
	\caption{\textbf{Comparison between existing tasks and ours.} (a) Video prediction methods focus on predicting future frames using historical video frames. (c) Motion prediction methods utilize a static map as an additional input, predicting future motion states based on the historical movement patterns of the target. (b) The proposed Target-Aware aerial Video Prediction task, which not only predicts the overall evolution of the environment but also focus on the motion state of target, achieving a more integrated spatiotemporal prediction.}
	\label{fig:intro}
\end{figure*}

Existing video prediction methods\cite{shi2015convolutional,shi2017deep,wang2018predrnn++,wang2022predrnn,gao2022simvp,tan2022simvp,tan2023temporal} typically explore spatiotemporal dynamics of historical consecutive frames to predict the future global scene (i.e., regressing pixel values for future frames), as shown in Fig. \ref{fig:intro}(a). However, they primarily focus on the changes in scene level, failing to explicitly capture the dynamic behavior of the target, especially when the target undergoes complex pose changes in the video sequence. Moreover, these methods often encounter the challenge of blurry predictions, where the output video frames are frequently unclear, increasingly worse when predicting further in the future\cite{mathieu2015deep}. This problem results in significant deformation of the target's appearance, making it difficult to identify appearance and positions of the target in future frames. The main reasons for generating blurry predictions lie in the complex spatiotemporal feature extraction and state transition operations inherent in these methods, leading to inevitable loss of appearance information. Additionally, the application of L2 loss exacerbates the blurriness of predictions. Despite numerous studies\cite{mathieu2015deep,tan2023temporal,zhong2023mmvp} attempting to maintain object appearance information while modeling scene evolution, the issue of blurry predictions persists, leading to the unresolved problem of target blurring.

There are also some research\cite{altche2017lstm,wiest2012probabilistic,nikhil2018convolutional,houenou2013vehicle,messaoud2020attention,DBLP:journals/corr/abs-2003-08111,mo2022multi}  that focus on the trajectory prediction of the target of interest, as shown in Fig. \ref{fig:intro}(c). They are dedicated to forecasting the motion trajectory of the targets, operating in lower data dimensions compared to video prediction. Nevertheless, they are unable to predict structural information in images, leading to a lack of foresight in the evolution of the global scene. While many studies\cite{yu2021dynamic,huang2020diversitygan} attempt to enhance models by incorporating the bird’s-eye view (BEV) image as an additional input to provide more environmental information, static BEV information may not offer sufficient contextual details, often lacking adaptability to environmental changes.

A more ideal solution involves unifying the modeling of environmental evolution and target motion, which allows the evolution information of the environment to guide motion prediction of the target, while target's motion information contributes to capture the environmental evolution. Therefore, we propose a novel task called Target-Aware Aerial Video Prediction, aiming to simultaneously predict future scenes and target's motion states, as shown in Fig. \ref{fig:intro}(b). In doing so, a better understanding and prediction of the changing trends in both global scene and local target can be achieved. The key challenge lies in aligning and interacting the information from different modalities, ensuring precise predictive capabilities for both global and local changes in aerial scenes. 

Further, we design a model specifically for this task named TAFormer, which is able to predict future video frames and target states by simultaneously considering historical video frames and target motion states. 
Specifically, we design a Target-Aware Transformer Encoder to unify the modeling of video and motion states of the target, which learns rich spatiotemporal dynamics in aerial videos and simultaneously captures motion patterns of the target. The Encoder employs a two-branch Transformer architecture to model both video and target motion, which is characterized by two ingenious designs: Spatiotemporal Attention (STA) and Information-Sharing Mechanism through Messengers (ISM). STA decouples the modeling of spatiotemporal dynamics in videos and applies spatial static attention and temporal dynamic attention separately to the video features, aimming to modeling object appearance and motion respectively. To facilitate the information interaction between video and motion states of the target, we design the ISM, which primarily consists of 3 parts: messengers initialization, message collecting, and message passing. Through the information-sharing mechanism, alignment and effective fusion of features from two modalities are achieved. The introduction of video (environmental) information aids in a more reasonable prediction of the target's future motion states, while the incorporation of motion information of the target supports a more precise target-sensitive prediction of environmental evolution. Moreover, to improve the model's performance in predicting the target's surrounding environment, we design Target-sensitive Gaussian Loss, which enforces the model to prioritize the accuracy and quality of predictions in the region around the target by applying increased penalty in the loss for this area. 

Due to the absence of the datasets specifically designed for target-aware video prediction, we adaptively modify the single object tracking (SOT) datasets UAV123\cite{benchmarkbenchmark} and VisDrone\cite{zhu2021detection} and conduct extensive experiments, demonstrating the outstanding performance of our model in Target-Aware Aerial Video Prediction. Overall, our work contributes in the following ways:
% To make the model's prediction sensitive to target's apperance and position, we design Target-Sensitive Gaussian Loss, which enforces the model to prioritize the accuracy and quality of predictions in the region around the target by applying increased penalty in the loss for this area.
\begin{enumerate}
    \item We propose a novel Target-Aware Aerial Video Prediction task, and design a model named TAFormer specifically for it, aiming to simultaneously predict future scenes and motion states of the target in aerial videos.
    \item  We design a Target-Aware Transformer Encoder to unify the modeling of video and motion states of the target, which fully explores the dynamics of video and target motion patterns, achieving interaction between scene information and target motion information through two sets of messenger tokens.
    \item To alleviate the difficulty of distinguishing targets in blurry predictions, we introduce Target-Sensitive Gaussian Loss (TSGL), enhancing the model's sensitivity to both target's position and content.
    \item  We adaptively modify the SOT datasets UAV123 and VisDrone and conduct extensive experiments, demonstrating the outstanding performance of the proposed TAformer in Target-Aware Aerial Video Prediction.
\end{enumerate}

\section{Related Work} \label{sec:related works}
This section primarily reviews previous research in video prediction, motion prediction and BEV-based future prediction for autonomous driving, analyzing their strengths and limitations.

\subsection{Video Predcition} \label{subsec:SPL}
Video prediction can be regarded as spatiotemporal predictive learning, which is a promising branch within the field of self-supervised learning\cite{liu2021self}, focusing on video-level information and predicting future video frames conditioned on historical frames. The development of deep learning\cite{dosovitskiy2020image,mao2022beyond,mao2022bidirectional,mao2023elevation,bi2023not,yao2023automated} has brought new dawn to video prediction.

Mainstream approaches\cite{shi2015convolutional,michalski2014modeling,wang2017predrnn,wang2018predrnn++,wang2022predrnn,wang2019memory,oliu2018folded} employ stacked RNNs for prediction. They intricately design structures to enhance the model's suitability for spatiotemporal prediction and alleviate challenges with modeling long-term dependencies. ConvLSTM\cite{shi2015convolutional} extends the fully connected LSTM (FC-LSTM) by incorporating convolutional structures in both the transitions from input to state and from state to state, enabling it to capture spatiotemporal correlations. PredRNN\cite{wang2017predrnn} employs spatiotemporal LSTM units to learn and memorize spatiotemporal representations. PredRNN++\cite{wang2018predrnn++} introduces a gradient highway unit that offers alternative fast paths for the gradient flow from the output to distant past inputs, enabling adaptive capture of both short and long term dependencies in videos. MIM\cite{wang2019memory} utilizes the differential signals between adjacent recursive states and employs two cascaded, self-updating memory modules to capture both the non-stationary and nearly stationary characteristics inherent in spatiotemporal dynamics. Given their outstanding adaptability and precision, these methods play a crucial role in video prediction.

Some studies attempt to abandon recurrent structures in pursuit of more efficient and accurate predictions. Past researches\cite{aigner2018futuregan,liu2017video} tend to utilize 3D convolutional networks to learn spatiotemporal dynamics. PredCNN\cite{xu2018predcnn} and TtrackCNN\cite{liu2020trajectorycnn} use 2D convolutional networks to enhance efficiency. However, these models exhibit lower prediction accuracy and fail to meet application requirements. Recently, SimVP\cite{tan2022simvp} has introduced a simple yet effective baseline model based on a pure CNN architecture, showcasing competitive performance. The team further proposes a video prediction framework OpenSTL\cite{tan2023openstl}, which extends MetaFormers\cite{yu2022metaformer} to bridge the gap between visual backbones and spatiotemporal prediction. This brings new insights to spatiotemporal predictive learning.

\subsection{Motion Prediction} \label{subsec:MP}
Motion prediction is a technology that involves forecasting the future positions of objects\cite{wu2020future}, people\cite{guo2023back}, or vehicles\cite{rezaei2023deep}, which relies on historical motion patterns and environmental factors to infer their likely trajectories. The applications of this field are extensive, including traffic management\cite{el2023deep}, autonomous driving\cite{rezaei2023deep} and video surveillance\cite{agarwal2023isgan}.

Presently, deep learning methods dominate this field, efficiently managing intricate scene information and attaining long-term predictions. Tpcn\cite{ye2021tpcn} argues that acquisition of temporal features is intricately linked to specified time intervals. It involves defining multiple time intervals, aggregating the agents' motion states within each interval, and subsequently conducting feature fusion across these intervals. To extract sufficient temporal information from agent movements, LaneGCN\cite{liang2020learning} employs 1D CNNs to encode motion state sequences along the temporal dimension and subsequently obtains and fuses multi-scale features. Some approaches\cite{ma2019trafficpredict,pan2020lane} consider temporal information by building spatiotemporal graphs or applying attention mechanisms along the temporal dimension. These methods\cite{kolekar2021behavior,song2022learning,messaoud2020attention,casas2020spagnn} frequently adhere to analogous paradigms: encoding the scene input to extract contextual features and subsequently decoding these features to generate predictions. Some studies\cite{cui2019multimodal,phan2020covernet,djuric2018short} utilize a BEV grid image to depict the surrounding scene information of targets, providing a more comprehensive spatial awareness. However, the static BEV image fails to capture the real-time status and changes of the environment and targets in dynamic scenes, still presenting shortcomings in addressing the issue of dynamic spatial perception.

\subsection{BEV-based Future Prediction for Autonomous Driving}
Precisely perceiving instances and forecasting their future motion is a critical task for autonomous vehicles\cite{li2023powerbev,zhai2023rethinking}, which is highly relevant to the proposed task of target-aware aerial video prediction. Early prediction methods based on BEV \cite{bansal2018chauffeurnet,hong2019rules} project past trajectories onto a BEV image. This approach assumes perfect detection and tracking of the target, utilizing CNNs to encode the rasterized input. Another category of works\cite{casas2018intentnet,luo2018fast,casas2020spagnn} involves direct end-to-end trajectory prediction from LiDAR point clouds. Unlike instance-level trajectory prediction, they handle the prediction task by assigning a motion (flow) field to every occupied grid. Differing from the previously mentioned methods that depend on LiDAR data, FIERY\cite{hu2021fiery} initially forecasts Bird's Eye View (BEV) instance segmentation exclusively from multi-view camera data. It learns from multiple frames in BEV after LSS\cite{philion2020lift} extraction, merges them into spatiotemporal states using recurrent networks, and subsequently performs probabilistic instance prediction. These methods follow Panoptic-DeepLab\cite{cheng2020panoptic}, which employs four distinct heads to compute semantic segmentation maps, instance centers, per-pixel centerness offset, and future flow. They depend on elaborate post-processing techniques to output the ultimate instance predictions from these four representations. In this paper, the target-aware aerial video prediction task we propose is addressed in an end-to-end manner, directly predicting future scenes and target motion states. There is no need for intricate post-processing, and the model is sensitive to changes in both the global scene and local target without requiring complex additional steps.

Indeed, BEV-based future prediction can be considered a paradigm of motion prediction, with its methods often relying on intricate post-processing steps and lacking the ability to anticipate the evolution of the global environment. In summary, both spatiotemporal predictive learning and motion prediction have their limitations. Spatiotemporal predictive learning may perform poorly in capturing specific target movements, while motion prediction might overlook the influence of environmental changes. The integration of the two tasks is expected to compensate for each other's shortcomings, offering a more robust solution for effective spatiotemporal understanding and motion prediction.

\section{Method}
This section provides a detailed exposition of the proposed method for target-aware aerial video prediction, named TAFormer. Section \ref{subsec:Problem Definition} outlines the defination for target-aware aerial video prediction. Section \ref{subsec:Overall Framework} provides an overall depiction of TAFormer. Section \ref{subsec:Spatiotemporal Attention} offers a detailed description of the Spatiotemporal Attention proposed for thorough exploration of video spatiotemporal dynamics. Section \ref{subsec:ISM} provides a comprehensive description of the Information Sharing Mechanism through Messengers between video and motion states. Finally, Section \ref{subsec:loss} details the Target-Sensitive Gaussian Loss used for model training.

\subsection{Problem Definition} \label{subsec:Problem Definition}
We formalize the definition of the target-aware aerial video prediction problem in the following manner. Provided a sequence of video frames $\mathcal{X}_{t,T}=\{\boldsymbol{x}_{i}\}_{t-T+1}^{t}$ and the corresponding bounding box sequence $\mathcal{B}_{t,T}=\{\boldsymbol{b}_{i}\}_{t-T+1}^{t}$ for the interested target, spanning the past $T$ frames up to time $t$, the goal is to forecast the succeeding $T'$ video frames $\mathcal{Y}_{t+1, T'}=\{\boldsymbol{x}_{i}\}_{t+1}^{t+1+T'}$ and bounding boxes for the target of interest $\mathcal{C}_{t+1, T'}=\{\boldsymbol{b}_{i}\}_{t+1}^{t+1+T'}$ commencing from time $t + 1$. Each video frame \(\boldsymbol{x}_i \in \mathbb{R}^{C \times H \times W}\) consists of $C$ channels, with height and width being $H$ and $W$, respectively. Each bounding box \(\boldsymbol{b}_i \in \mathbb{R}^4\), where the 4 elements represent the center coordinates and the width and height of the bounding box. In practical scenarios, we denote the input frames and the predicted frames as tensors $\mathcal{X}_{t,T} \in \mathbb{R}^{T \times C \times H \times W}$ and $\mathcal{Y}_{t+1,T'} \in \mathbb{R}^{T' \times C \times H \times W}$, and represent the input bounding box sequence and the output bounding box sequence as tensors $\mathcal{B}_{t,T} \in \mathbb{R}^{T \times 4}$ and $\mathcal{C}_{t+1,T'} \in \mathbb{R}^{T' \times 4}$, respectively.

The model with trainable parameters $\theta$ learns a mapping $\mathcal{F}_{\theta}: (\mathcal{X}_{t,T}, \mathcal{B}_{t,T}) \mapsto (\mathcal{Y}_{t+1,T'}, \mathcal{C}_{t+1,T'})$ by simultaneously exploring the spatiotemporal dynamics of the video and the motion patterns of the interested target. In this paper, $\mathcal{F}_{\theta}$ corresponds to a trained neural network, aiming to minimize the differences between the predicted future frames and the actual future frames, as well as the differences between the predicted future positions of the target and the actual future positions of the target. The optimal model parameters can be represented as:
\begin{equation}
\theta^{\ast}=\mathop{\arg\min}\limits_{\theta}\left(\mathcal{L}\left(\operatorname{\mathcal{F}_{\theta}}\left(\mathcal{X}_{t,T}, \mathcal{B}_{t,T}\right),\mathcal{Y}_{t+1,T'}, \mathcal{C}_{t+1,T'}\right)\right)
    \label{optim_Eq}
\end{equation}
where $\mathcal{L}$ represents the loss function, quantifying the disparities between the predicted and actual future frames, as well as the disparities between the forecasted and true future positions of the target.

\subsection{Overview} \label{subsec:Overall Framework}
\begin{figure*}[htbp]
	\setlength{\abovecaptionskip}{1pt}
	\centering
	\includegraphics[width=0.8\linewidth]{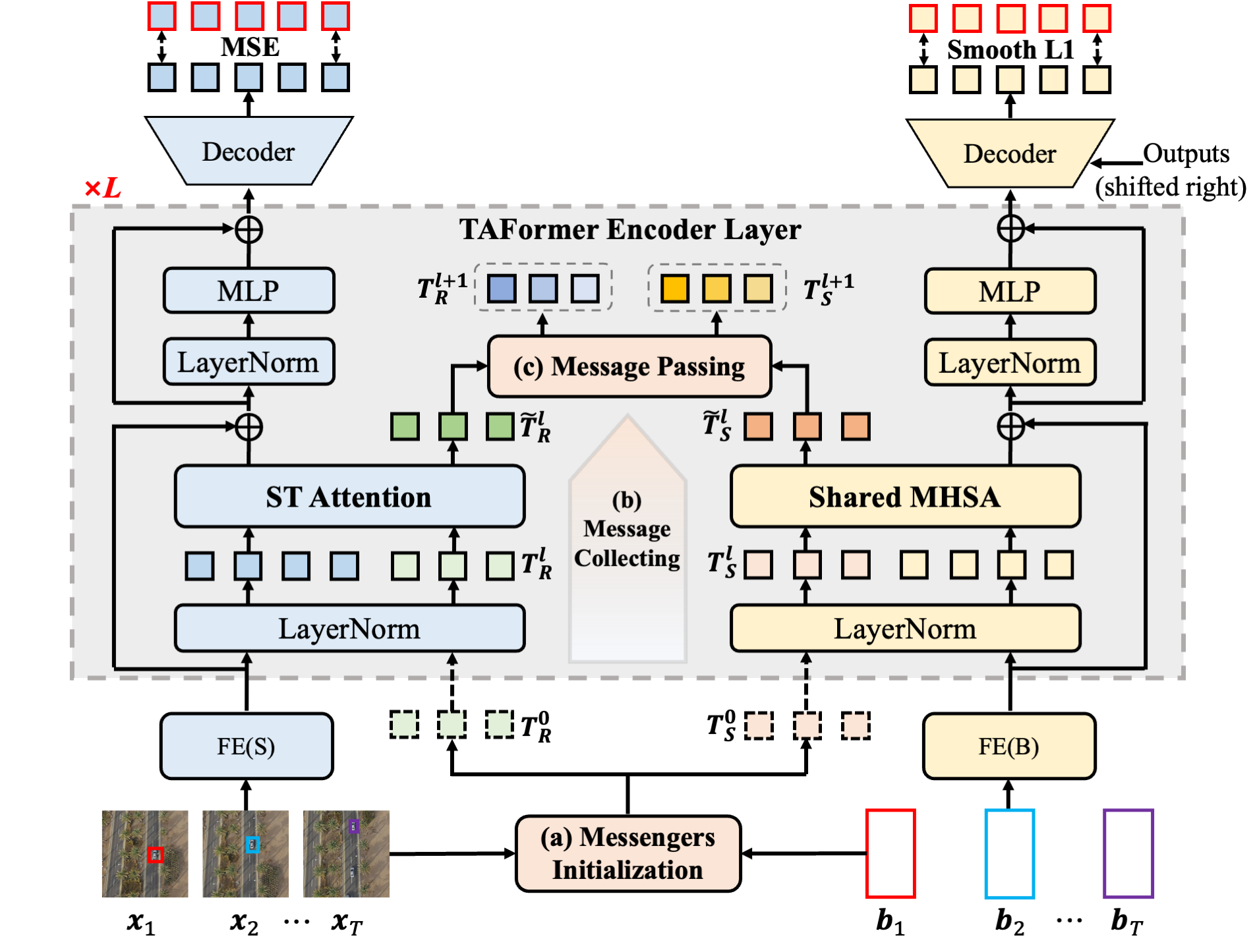}
	\caption{\textbf{The overall framework of TAFormer. }Provided a sequence of video frames $\mathcal{X}_{t,T}=\{\boldsymbol{x}_{i}\}_{t-T+1}^{t}$ and the corresponding bounding box sequence $\mathcal{B}_{t,T}=\{\boldsymbol{b}_{i}\}_{t-T+1}^{t}$ for the interested target, spanning the past $T$ frames up to time $t$, TAFormer is capable to forecast the succeeding $T'$ video frames $\mathcal{Y}_{t+1, T'}=\{\boldsymbol{x}_{i}\}_{t+1}^{t+1+T'}$ and bounding boxes for the target of interest $\mathcal{C}_{t+1, T'}=\{\boldsymbol{b}_{i}\}_{t+1}^{t+1+T'}$ commencing from time $t + 1$. FE(S) and FE(B) represent spatial feature embedding and bounding box feature embedding, respectively.}
	\label{fig:TAformer}
\end{figure*}
To accurately predict future frames of aerial videos and the future states of the interested target, we integrate spatiotemporal predictive learning with motion prediction and propose a method for target-aware aerial video prediction, named TAFormer. The overall structure is illustrated in Fig. \ref{fig:TAformer}.  The main components of TAFormer include Feature Embedding, Target-Aware Transformer Encoder and Decoders, as well as the loss functions.

\noindent \textbf{Feature Embedding\quad}The historical video frames \(\mathcal{X}_{t,T}=\{\boldsymbol{x}_{i}\}_{t-T+1}^{t}\) pass through the Spatial Feature Embedding (FE(S) in Fig. \ref{fig:TAformer}), consisting of four vanilla 2D convolutional layers, resulting in a concise spatial feature representation \(Z = \{{\boldsymbol{z}_i\}}_{t-T+1}^t\). Simultaneously, the historical bounding boxes \(\mathcal{B}_{t,T}=\{\boldsymbol{b}_{i}\}_{t-T+1}^{t}\) of the target undergo a linear layer (FE(S) in Fig. \ref{fig:TAformer}) and position encoding, as per \cite{DBLP:journals/corr/abs-2003-08111}, yielding the motion states embedding \(S = \{{\boldsymbol{s}_i\}}_{t-T+1}^t\).

\noindent \textbf{Target-Aware Transformer Encoder\quad}Subsequently, \(Z\) and \(S\) are fed into the core of the model: Target-Aware Transformer Encoder (TA Encoder), which is crafted to learn spatiotemporal dynamics in aerial videos while capturing the historical motion patterns of the interested target. For the spatial dense representation \(Z \in \mathbb{R}^{T \times C_{hid} \times h \times w}\) of the historical video frame sequence, we compress the time and channel dimensions into a unified dimension, treating the features at each spatial position as a token, resulting in \(F \in \mathbb{R}^{(h \times w) \times C'}\), where $C'=T \times C_{hid}$. Subsequently, we apply Spatiotemporal Attention and multi-layer perceptron (MLP) to learn the spatiotemporal dynamics:
\begin{equation}{\hat{F}}^l=\operatorname{STA}\left(\operatorname{LN}\left(F^{l-1}\right)\right)+F^{l-1}
\end{equation}
\begin{equation}
    F^l = \operatorname{MLP}(\operatorname{LN}({\hat{F}}^l))+{\hat{F}}^l
\end{equation}
where $F^l$ represents the output of the $l$-th layer of the TA Encoder, $\operatorname{LN}(\cdot)$ represents layer normalization\cite{ba2016layer}, following the transformer architecture, $\operatorname{STA}$ is short for Spatiotemporal Attention. Similarly, we employ multi-head self-attention (MHSA) and MLP to mine the historical motion patterns in the motion state embedding $S \in  \mathbb{R}^{T \times C'}$:
\begin{equation} \label{eq:4}
{\hat{S}}^l=\operatorname{MHSA}\left(\operatorname{LN}\left(S^{l-1}\right)\right)+S^{l-1}
\end{equation}
\begin{equation}
    S^l = \operatorname{MLP}(\operatorname{LN}({\hat{S}}^l))+{\hat{S}}^l
\end{equation}
where $S^l$ represents the output of the TA Encoder at layer $l$. To further facilitate the interaction between video and motion information of the target, we design the Information-Sharing Mechanism through Messengers (ISM), which comprises three components: messengers initialization, message collecting, and message passing. The messengers initialization phase is conducted prior to the TA Encoder, initializing ROI tokens and state tokens. Subsequent message learning and message passing stages occur internally within the TA Encoder. During message learning, ROI tokens gather information around the regions of interest in the video, and state tokens collect motion state information of the target of interest. In the message passing stage, messenger tokens interact with each other, transmitting information to the next layer. Through the information-sharing mechanism, alignment and effective fusion of features from two modalities are achieved. 

\noindent \textbf{Decoder\quad}The encoded video features are separated along the channel-time dimensions to obtain \(\widetilde{F}\in\mathbb{R}^{T\times C_{hid}\times h\times w}\), followed by the video decoder, consisting of four layers of ConvTranspose2d\cite{dumoulin2016guide}, to produce the final vodeo prediction \(\hat{Y}\in\mathbb{R}^{T\times C\times H\times W}\). The encoded historical motion state features are passed through the motion decoder (four vanilla Transformer Decoder layers following the classical Transformer architecture\cite{vaswani2017attention}) to perform autoregressive prediction for the future bounding boxes of the target, resulting in \(\hat{C} \in \mathbb{R}^{T' \times 4}\).

\noindent \textbf{Loss Functions\quad}
We employ mean squared error (MSE) loss to measure the difference between predicted video frames and actual video frames, and use Smooth L1 loss to quantify the difference between predicted bounding boxes and actual bounding boxes.
\begin{equation}
    \mathcal{L}_{video} = \frac{1}{T^{'}} \sum_{t+1}^{t+T^{'}}\left(\hat{\boldsymbol{y}}^{i} - \boldsymbol{y}^{i}\right)^{2}
\end{equation}
\begin{equation}
    \mathcal{L}_{motion} =  \frac{1}{T^{'}} \sum_{t+1}^{t+T^{'}} \begin{cases}
       \frac{1}{2} \times \left(\hat{\boldsymbol{c}}^{i} - \boldsymbol{c}^{i}\right)^{2} , & |\hat{\boldsymbol{c}}^{i} - \boldsymbol{c}^{i}| < 1 \\
       |\hat{\boldsymbol{c}}^{i} - \boldsymbol{c}^{i}| - \frac{1}{2} , & |\hat{\boldsymbol{c}}^{i} - \boldsymbol{c}^{i}| \ge 1
    \end{cases}
\end{equation}
To enhance the model’s accuracy in predicting the target’s surrounding environment, we design Target-Sensitive Gaussian Loss, denoted as $\mathcal{L}_{Gaussian}$, which enforces the model to prioritize the accuracy and quality of predictions in the region around the target by applying increased penalty in the loss for this area. The comprehensive loss function of the model is obtained by the weighted sum of the three losses:
\begin{equation}
    \mathcal{L} = \mathcal{L}_{video} + \lambda_1 \cdot \mathcal{L}_{moiton} + \lambda_2 \cdot \mathcal{L}_{Gaussian}
\end{equation}
where $\lambda_1$ and $\lambda_2$ are the weights controlling the three losses.

The remaining part of this section describes three key designs of the proposed method: Spatiotemporal Attention, Information-Sharing Mechanism through Messengers, and Target-Sensitive Gaussian Loss.

\subsection{Spatiotemporal Attention} \label{subsec:Spatiotemporal Attention}
\begin{figure}[t]
	\setlength{\abovecaptionskip}{1pt}
	\centering
	\includegraphics[width=1.0\linewidth]{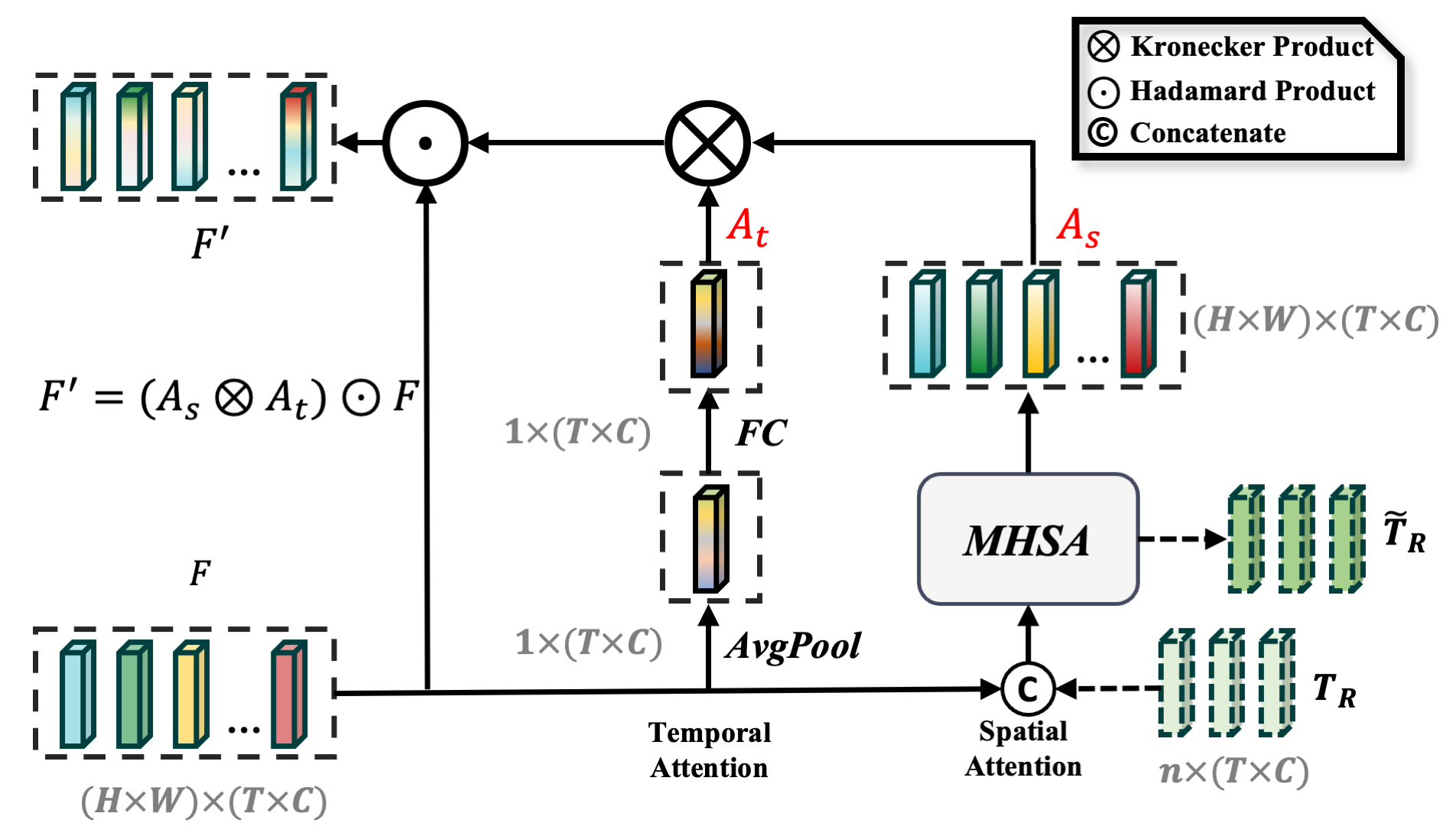}
	\caption{\textbf{Details of the Spatiotemporal Attention.} It consists of spatial attention and temporal attention, and the final attention is the product of them.}
	\label{fig:STA}
\end{figure}

When processing video input, both the spatial feature embedding and the video decoder of the model only perform convolutions or transposed convolutions in the spatial dimensions, without modeling the temporal correlations among video frames. To further explore the spatiotemporal dynamics in aerial videos and model the scene appearance and motion, we design Spatiotemporal Attention, as illustrated in Fig. \ref{fig:STA}.

We reshape the spatially embedded features to obtain \(F \in \mathbb{R}^{(H \times W) \times (T \times C_{hid})}\), arranging the frames sequentially along the channel dimension. {Inspired by the widespread success of vision Transformers (ViTs)\cite{dosovitskiy2020image,liu2021swin} and aiming to unify the modeling of both the videos and target motion states, we treat the vector at each spatial position in the feature map as a token and apply multi-head self-attention (MHSA) to them, allowing each spatial position to attend to all others, significantly expanding the receptive field.} MHSA effectively models long-range spatial correlations by computing the similarity of tokens at different spatial positions, thus better capturing scene appearance information. 

However, relying solely on MHSA is insufficient for fully exploring temporal evolution patterns and lacks modeling of the motion of the scene and objects within it. In order to more comprehensively learn the temporal evolution between frames, we introduce a temporal attention mechanism that dynamically learns attention weights for channels, adopting a squeeze-and-excitation (SE) approach\cite{hu2018squeeze} { due to its simple, direct, and efficient nature. The SE mechanism learns the correlations between channels and weights each channel based on its importance, effectively extracting features and enhancing the model's representational capacity. This mechanism does not require complex network structures or a large number of parameters, making its implementation relatively simple, and it can improve model performance without adding too much computational cost.} The final attention is the Kronecker product of spatial attention and temporal attention.

It is worth noting that the operations connected by dashed arrows in Fig. \ref{fig:STA} are part of the information collecting stage in ISM (detailed in Section \ref{subsec:ISM}). We concatenate ROI messenger tokens $T_R$ initialized with features from the surrounding region of the target and video tokens, and input them together into MHSA. This is a clever design that achieves two goals at once: through self-attention, ROI messenger tokens gather information about the target's surrounding area in the video, and video tokens pay more attention to the region around the target,  making the appearance modeling of the region around the target more reliable.

The proposed spatiotemporal attention can be formalized as follows:
\begin{equation}
\label{eq:9}
    A_{s}, \tilde{T}_R = \operatorname{MHSA}(\operatorname{LN}([F, T_R])
\end{equation}
\begin{equation}
    A_{t} = \operatorname{FC}\left(\operatorname{AvgPool}\left(F\right)\right)
\end{equation}
\begin{equation}
    F' = ( A_{t} \otimes A_{s}) \odot F  
\end{equation}
where $[\text{·}]$ denotes the concatenation operation, \(A_s\) represents the result of spatial attention, \(A_t\) represents the result of temporal attention, \(\operatorname{FC}\) is a fully connected layer, and \(AvgPool\) is an average pooling layer. $\otimes$ and $\odot$ denote the Kronecker Product and the Hadamard Product, respectively.

\subsection{Information Sharing Mechanism through Messengers} 
\label{subsec:ISM}
\begin{figure}[t]
	\setlength{\abovecaptionskip}{1pt}
	\centering
	\includegraphics[width=1.0\linewidth]{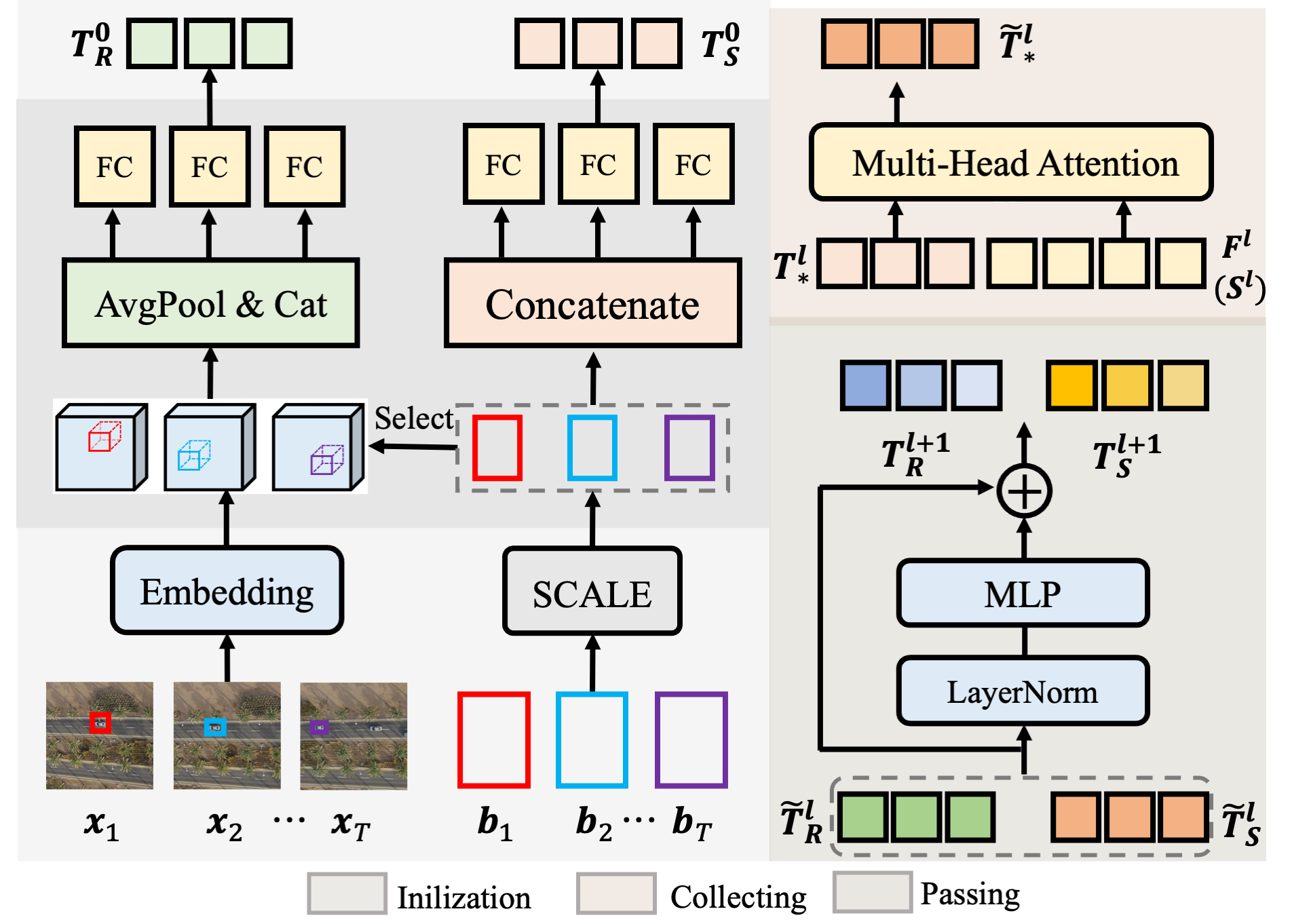}
	\caption{\textbf{Details of the ISM.} Different background colors represent distinct processing steps. The left half illustrates the Messengers Initialization process, while the upper and lower sections on the right depict the Message Collecting and Message Passing processes, respectively.}
	\label{fig:ISM}
\end{figure}

To provide essential environmental information for predicting the motion states of the target and simultaneously leverage the target's motion information to enhance the accuracy of future video frames prediction, we devise an Information-Sharing Mechanism through Messengers (ISM), which comprises three main components: Messengers Initialization, Message Collecting, and Message Passing, as illustrated in Fig. \ref{fig:ISM}. Through the information-sharing mechanism, features from both modalities are aligned, facilitating effective fusion. The introduction of environmental information aids in the precise and rational prediction of future motion states, while the inclusion of historical state information contributes to a more accurate and target-sensitive forecast of environmental evolution.
\subsubsection{Messenger Initialization} \label{subsubsec:init}
The mission of the messengers is to transmit information between video features and target motion features. To achieve this, we initialize two types of messengers: region of interest (ROI) tokens (denoted as $T_R$) and state tokens (denoted as $T_S$). The specific process is illustrated in the left part of Fig. \ref{fig:ISM}. 

The ROI messengers represents the environmental features around the target of interest. In Section \ref{subsec:Overall Framework}, we apply spatial feature embedding to each frame $\boldsymbol{x}_{i}$  in the historical video sequence and obtain a compact representation of spatial feature \(\boldsymbol{z}_i\) (downsampled by a factor of 4). Here, we scale the bounding box \(\boldsymbol{b}_{i}\) corresponding to the respective frame to match the current feature map's size and get \(\boldsymbol{b}_{i}'\). Subsequently, we select the corresponding region of interest on each frame's feature map, obtain the features of the regions of interest through average pooling, 
concatenate them along the temporal dimension, and finally use linear transformations to get $M$ different tokens.
\begin{equation}    
\boldsymbol{r}_i = \operatorname{AvgPool}\left(\operatorname{Select}\left(\operatorname{Scale}\left(\boldsymbol{b}_i\right), \boldsymbol{z}_i\right) \right)
\end{equation}
\begin{equation}    
\boldsymbol{t}_r^m =\operatorname{FC}_{m}\left([\{\boldsymbol{r}_{i}\}_{t-T+1}^{T}] \right)
\end{equation}
\begin{equation}    
T_R = \{\boldsymbol{t}_r^m\}_{m=1}^{M}
\end{equation}
where $b_i \in \mathbb{R}^{T \times 4}$, $\boldsymbol{t}_r^m \in \mathbb{R}^{C'}$ represents the $m$th ROI token, $\operatorname{Select}(\cdot, \cdot)$ means to select the vector in the feature map corresponding to the coordinates, and $\operatorname{FC}$ is short for fully connection.

The State messengers carry historical motion state information of the target. The scaled bounding box vectors are concatenated along the temporal dimension and then passed through several fully connected layers to obtain $N$ state tokens.
\begin{equation}
    \boldsymbol{t}_{s}^{n} = \operatorname{FC}_{n}\left([\operatorname{Scale}(\{\boldsymbol{b}_{i}\}_{t-T+1}^{T})]\right)
\end{equation}
\begin{equation}    
T_S = \{\boldsymbol{t}_s^n\}_{n=1}^{N}
\end{equation}
where $\boldsymbol{t}_s^n \in \mathbb{R}^{C'}$ denote the $n$th state token.

\subsubsection{Message Collecting}
The initialized messenger tokens carry fundamental information for their respective missions. To achieve the ultimate goal of information sharing, two essential tasks must be accomplished: firstly, enriching the environmental information carried by $T_R$ and making the target motion state information represented by $T_S$ more comprehensive; secondly, allowing $T_R$ to teach the learned target state information to the video tokens after interacting with $T_S$ during message passing, and vice versa for $T_S$ teaching environmental information to the motion branch tokens. Both of the two can be accomplished through MHSA, as illustrated in the upper-right part of Fig. \ref{fig:ISM}. The capability of MHSA to model complex dependencies across different tokens in a sequence makes it an ideal mechanism for achieving both tasks efficiently, as it allows the model to selectively focus on relevant information within the messenger tokens and effectively capture the intricate relationships between the messenger tokens and the video (or motion) tokens.
% \begin{equation}
%     A_{s}, \tilde{T}_R^{l} = \operatorname{MHSA}(\operatorname{LN}([F, T_R^{l-1}])))
% \end{equation}
% \begin{equation}
% {\tilde{S}}^{l}, \tilde{T}_S^{l}=\operatorname{MHSA}\left(\operatorname{LN}\left([S^{l-1}, T_S^{l-1}]\right)\right)
% \end{equation}
% where $l$ denotes the $l$th layer of the TAFormer encoder.
\subsubsection{Message Passing}
After Message Learning, $T_R$ acquires richer environmental information related to the target, while $T_S$ carries more comprehensive information about target motion. In the message passing stage, to facilitate the interaction and sharing of information between the two messengers, we employ a simple multi-layer perceptron (MLP):
\begin{equation}
T_R^l, T_S^l = \operatorname{MLP}(\operatorname{LN}([\tilde{T}_R^l, \tilde{T}_S^l])) + [\tilde{T}_R^l, \tilde{T}_S^l]
\end{equation}
where $T_R^l$, $T_S^l$ represent the messenger tokens entering the next layer of the TAFormer Encoder. In MLP, each neuron in every layer is connected to all neurons in the preceding layer. This operation enables the features of \(T_R^l\) and \(T_S^l\) to interact through a linear combination with the weight matrix. The introduction of a non-linear activation function in the MLP introduces non-linear transformations, thereby enhancing the expressive power of features. This design allows for complex information exchange and interaction between messenger tokens, aiding in better alignment and fusion of features from the two modalities.
\subsection{Target-Sensitive Gaussian Loss} \label{subsec:loss}
{In traditional video prediction methods, the mean squared error (MSE) between predicted frames and ground truth is frequently employed as the loss function to guide model training, with the objective of minimizing the discrepancy between generated predictions and actual observations. In such training processes, the model treats each pixel in the video frames equally. However, this training paradigm suffers from blurred predictions, especially as the prediction time extends, making it difficult to distinguish between targets and backgrounds in the predicted frames. Moreover, in aerial videos, targets typically appear small in scale, exacerbating the difficulty of distinguishing targets within the predicted frames.} 

\begin{figure}[t]
	\setlength{\abovecaptionskip}{1pt}
	\centering
	\includegraphics[width=1.0\linewidth]{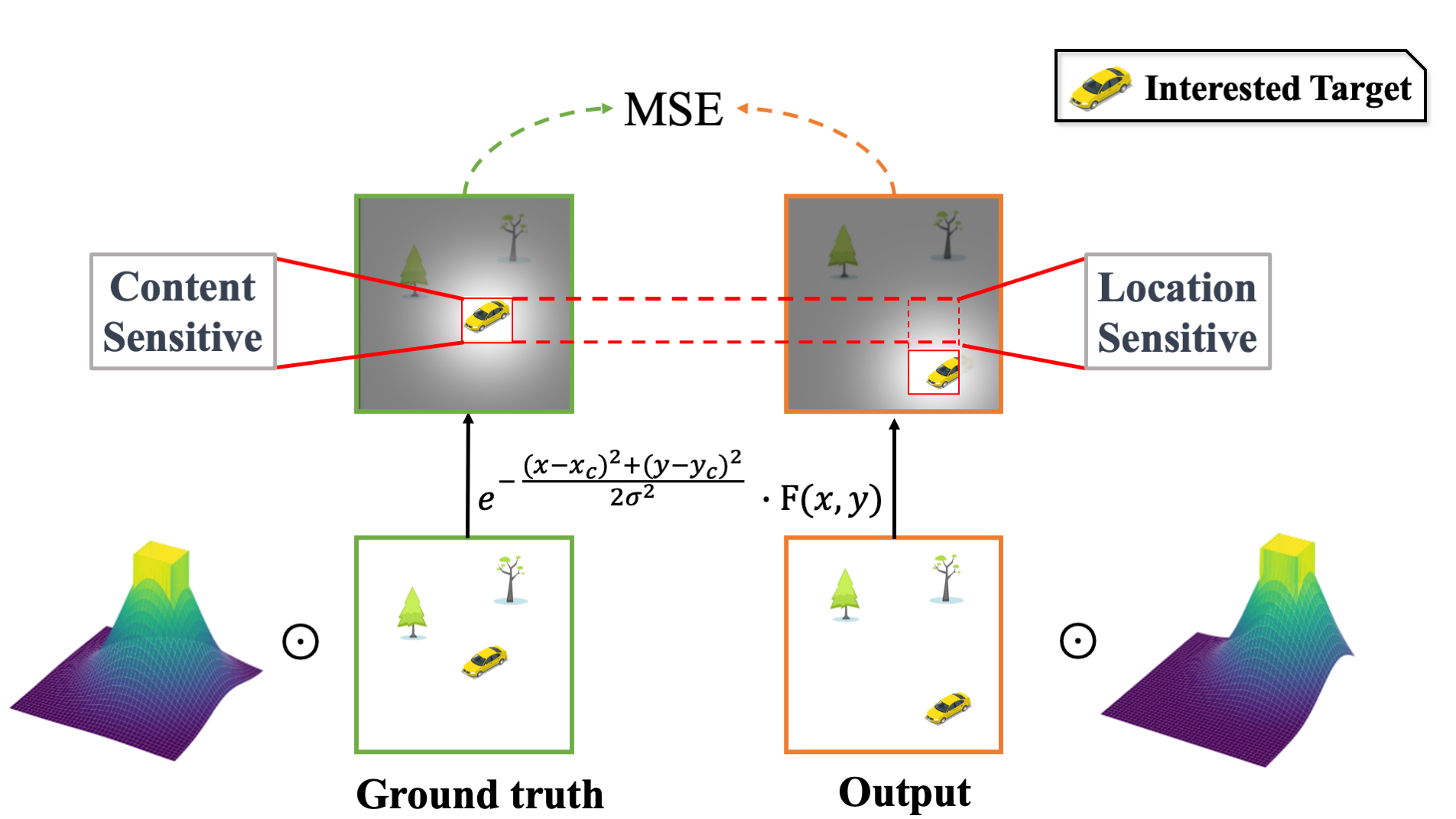}
	\caption{\textbf{Target-sensitive Gaussian loss.} By applying Gaussian weighting to the predicted frames and ground truth, we highlight the regions relevant to the interested target, achieving content-awareness and position-awareness for the target.}
	\label{fig:loss}
\end{figure}

{To address this issue, an intuitive approach is to weight different spatial regions of the predicted frame and the ground truth before computing the loss, thereby highlighting the target. We firmly believe that regions closer to the target are more crucial for modeling its position and appearance. The spatial distribution properties of the two-dimensional Gaussian function make it suitable for assigning weights based on the distance from a point to the target center. Thus, we introduce the Target-Sensitive Gaussian Loss, which emphasizes the target by applying Gaussian weighting to both predicted and ground truth values. This method enhances both content and positional awareness for the target of interest, as depicted in Fig. \ref{fig:loss}.}

Specifically, here is the representation of a two-dimensional Gaussian function used for weighting:
% \begin{equation}
%  w^{\star}_{x, y} = \exp\left(-\frac{1}{2}\left(\frac{(x - x_c^\star)^2}{\sigma_x^2} + \frac{(y - y_c^\star)^2}{\sigma_y^2}\right)\right) 
% \end{equation}
\begin{equation}
    w^{\star}_{x, y}  =  \begin{cases}
    1 , \quad |x - x_c^\star| < \frac{w^\star}{2}\quad and \quad |y - y_c^\star| < \frac{h^\star}{2}\\
    \\
       \exp\left(-\frac{1}{2}\left(\frac{(x - x_c^\star)^2}{\sigma_x^2} + \frac{(y - y_c^\star)^2}{\sigma_y^2}\right)\right),  otherwise
    \end{cases}
\end{equation}
where \( (x, y) \) represents the spatial coordinates, $\star \in (P, G)$ denotes prediction or ground truth, \( (x_c^\star, y_c^\star, w^{\star}, h^{\star}) \) are the coordinates of the center point, width and height of the bounding box, and \( \sigma_x \) and \( \sigma_y \) are the standard deviations. In practice, we set $\sigma_x=\sigma_y=50$. This function assigns weights to different spatial positions outside the bounding box based on their distance from the center of it and assign weights 1 to the positions inside the bounding box. We denote the weight matrix for $\star \in (P, G)$ as:
Certainly, here is the adjusted equation with reduced width and \( m \) replaced by \( w \):
\begin{equation}
\tilde{W}^{\star} = \left(\begin{array}{cccc}
w^{\star}_{0,0} & w^{\star}_{0,1} & \cdots & w^{\star}_{0, W-1} \\
\vdots & & & \vdots \\
w^{\star}_{H-1,0} & w^{\star}_{H-1,1} & \cdots & w^{\star}_{H-1, W-1}
\end{array}\right)
\end{equation}
where $H$ and $W$ represent the height and width of the frames. The final target-sensitive Gaussian loss is computed as follows:
\begin{equation}
    \mathcal{L}_{Gaussian} = \frac{1}{T^{'}} \sum_{t+1}^{t+T^{'}}\left(\tilde{W}^{P} \odot \hat{\boldsymbol{y}}^{i} - \tilde{W}^{G} \odot \boldsymbol{y}^{i}\right)^{2}
\end{equation}
where $\hat{\boldsymbol{y}}^{i}$ and $\boldsymbol{y}^{i}$ represent the $i$th frame of the predicted and ground truth sequence, respectively.

\section{Experiments}
We perform comprehensive experiments to showcase the performance of the proposed model in target-aware video prediction. Section \ref{exp:set} introduces the experimental setup, including details about the datasets, implementation details, and evaluation metrics for model performance. Subsequently, Section \ref{exp:com} compares our proposed method with the current state-of-the-art approaches, showcasing the superiority of our model in predicting future scene evolution and target states. Section \ref{exp:abl} provides an overview of ablation experiments, evaluating the contributions of each proposed module and experimenting with various hyperparameter configurations. Finally, Section \ref{exp:ana} analyzes specific cases, including rapidly changing scenes, distance-sensitive target size variations, small targets, and examines instances where the model may fail.

\subsection{Experimental Setting} \label{exp:set}
\subsubsection{Datasets} \label{exp:data}
We adapt two SOT datasets, UAV123\cite{benchmarkbenchmark} and VisDrone\cite{zhu2021detection}, for target-aware video prediction, resulting in UAV123VP and VisDroneVP, respectively.

\noindent \textbf{UAV123VP} \quad The UAV123 dataset is a SOT dataset captured by low-altitude UAVs. It comprises a total of 123 video sequences with over 110K frames, featuring a resolution of 1280$\times$720 pixels. All sequences are fully annotated with vertical bounding boxes. The dataset encompasses diverse scenes, various types of targets, and actions. Additionally, the camera perspective continuously changes with the drone's movement, leading to significant variations in the size and aspect ratio of the bounding boxes relative to the initial frame, thereby increasing the difficulty of prediction. To adapt the dataset for the task of target-aware video prediction, we utilize a sliding window with a width of 16 to sample each video sequence, using a stride of 6, and correspondingly sample the annotation information. We partition the dataset into training, validation, and test sets with a ratio of 7:2:1, ensuring that samples within each set do not come from the same video sequence to prevent information leakage. This resulted in a total of 7106 samples for training, 2043 samples for validation, and 1023 samples for testing.

\noindent \textbf{VisDroneVP} \quad The VisDrone dataset consists of 288 video clips collected using different drone platforms in various scenes, weather conditions, and lighting conditions, with varying resolutions and wide coverage. The dataset includes rich annotations suitable for tasks such as object detection, single-object tracking, and crowd counting in both images and videos. To make it suitable for target-aware video prediction, we transformed the VisDrone-SOT dataset, originally used for single-object tracking. Since the original dataset's test set lacked annotations, we only utilized video segments from the training and validation sets. For each video segment, we sample using a sliding window with a width of 16 and a stride of 16. This resulted in 2146 samples sampled from the original training set, with 2000 used for training and 146 for validation. Additionally, we sample 403 samples from the original validation set for testing in the context of target-aware video prediction.
\subsubsection{Implementation Details} \label{exp:imp}
The proposed method is implemented using the PyTorch\cite{paszke2019pytorch} deep learning framework on a single NVIDIA RTX A40 GPU. The model is trained end-to-end using the Adam optimizer, with the momentum set to 0.9, and a learning rate of 0.001. Additionally, we employ the OneCycle strategy to dynamically adjust the learning rate during the training process. For experiments on both datasets, we resize the videos to a resolution of 256$\times$256 (the bounding box annotations of the target are also scaled proportionally). The batch size is set to 4, and the model is trained for 50 epochs. To assess the model's performance under different lengths of observed frames, we conduct experiments with two configurations for each dataset: predicting the future 8 frames using 8 observed frames (8→8) and predicting the future 8 frames using 4 observed frames (4→8).

The video spatial embedding module, video decoder, and motion decoder of the model all have 4 layers. The depth of the TAFormer Encoder is set to 6, and the numbers of ROI tokens and state tokens are set to 8 and 2, respectively. The Encoder and Decoder follow a wide-narrow design, with an embedding dimension of 512 for the Encoder and 64 for both Decoders. For the hyperparameters controlling the loss weights, $\lambda_1$ and $\lambda_2$ are both set to 0.001.

As there is currently no research applying video prediction in the remote sensing domain, and there is even less research on jointly predicting the motion of future video frames and the motion of the target of interest, we select recent outstanding methods in the computer vision (CV) domain for comparison. To ensure fairness, we use the same number of training epochs, batch size, and optimization strategy for all methods.
\subsubsection{Evaluation Metrics} \label{exp:eva}
For the assessment of future frame predictions, following OpenSTL\cite{tan2023openstl}, we comprehensively and rigorously assessed the model's performance using the following metrics:

\noindent \textbf{MSE and MAE.} We empoly MSE and mean absolute error (MAE) to measure the difference between the predicted results and the ground truth. 
% \begin{equation}
%    \text{MSE} = \frac{1}{T'} \sum_{i=1}^{T'} (y_i - \hat{y}_i)^2, \quad
%     \text{MAE} = \frac{1}{T'} \sum_{i=1}^{T'} |y_i - \hat{y}_i|
% \end{equation}
% where \(T'\) is the total number of frames, \(y_i\) is the ground truth, and \(\hat{y}_i\) is the predicted value for the \(i\)-th frame.

\noindent \textbf{SSIM and PSNR.} We use Structural Similarity Index (SSIM) and Peak Signal-to-Noise Ratio (PSNR) to measure the similarity between predicted results and the true target. They find extensive applications in image processing and computer vision.
% \begin{equation}
%     \text{SSIM} = \frac{1}{T'} \sum_{i=1}^{T'} \frac{{(2\mu_{y_i}\mu_{\hat{y}_i} + C_1)(2\sigma_{y_{i}\hat{y}_{i}} + C_2)}}{{(\mu_{y_i}^2 + \mu_{\hat{y}_{i}}^2 + C_1)(\sigma_{y_{i}}^2 + \sigma_{\hat{y}
% _{i}}^2 + C_2)}} 
% \end{equation}
% \begin{equation}
%    \text{PSNR} = \frac{1}{T'} \sum_{i=1}^{T'} 10 \cdot \log_{10}\left(\frac{{\text{MAX}^2}}{{(y_i - \hat{y}_i)^2}}\right)
% \end{equation}
% where \(y_i\) and \(\hat{y}_i\) represent the predicted and true target of the i-th frame, respectively. \(\mu\) is the mean, \(\sigma\) is the standard deviation, \(\sigma_{y_{i}\hat{y}_{i}}\) is the covariance, and \(C_1\) and \(C_2\) are constants for numerical stability. \(\text{MAX}\) is the maximum possible pixel value. 

\noindent \textbf{LPIPS.} Learned Perceptual Image Patch Similarity (LPIPS) provides a perceptual alignment evaluation method for visual tasks, assessing the perceptual difference between predicted result and target in the human visual system.

\noindent \textbf{Flops and FPS.} We employ the count of floating-point operations (Flops) as a metric to gauge the computational complexity of the model. Furthermore, we provide the frames per second (FPS) on the NVIDIA RTX A40 to assess the inference speed.

For the accuracy of predicting the future state of the target, we evaluate the model's performance using the following metrics:

\noindent \textbf{ROI-MSE.} In target-aware video prediction, we aim for the model to focus on predicting the region of interest, which includes the target and its surrounding area. To assess the quality of predictions in this specific region, we employ the target bounding box to select the region of interest and calculate the mean squared error loss, referred to as ROI-MSE.
\begin{equation}
    \text{ROI-MSE} = \frac{1}{T'} \sum_{i=1}^{T'} \| y_{r, i} - \hat{y}_{r, i} \|^2 
\end{equation}
where $y_{r, i}$ and $\hat{y}_{r, i}$ represent the region of interest in the predicted and true values of the i-th frame, respectively.

\noindent \textbf{mIoU and ADE.} We evaluate the model's accuracy in predicting the future motion states of the target by using the average intersection over union (mIoU) of predicted and true values of the region of interest bounding boxes in the sequence, as well as the average displacement error (ADE) of predicted and true values of the target box center points. The formulas are given by:
\begin{equation}
    \text{mIoU} = \frac{1}{T'} \sum_{i=1}^{T'} \frac{{Area(\boldsymbol{c}_i \cap \hat{\boldsymbol{c}}_i)}}{{Area(\boldsymbol{c}_i \cup \hat{\boldsymbol{c}}_i)}}
\end{equation}
\begin{equation}
 \text{ADE} = \frac{1}{T'} \sum_{i=1}^{T'} \sqrt{{(x_{c,i}-\hat{x}_{c,i})^2 + (y_{c,i}-\hat{y}_{c,i})^2}}
\end{equation}
where T' is the total number of frames in the sequence, and \(Area(\cdot)\) denotes the area of the corresponding bounding box,\((x_{c,i}, y_{c,i})\) and \((\hat{x}_{c,i}, \hat{y}_{c,i})\) denote the center point coordinates of the target bounding box for the i-th frame in the prediction and ground truth, respectively.

\subsection{Comparison with State-of-the-arts} \label{exp:com}
{In this paper, we introduce the task of target-aware video prediction and propose TAFormer for the parallel prediction of future scenes and the target's motion states. Interestingly, there exists a sequential solution to achieve the unified prediction of future scenes and the states of the target which involves tracking the target in the predicted frames, thereby estimating the future motion states.} 

{In this section, we not only compare TAFormer with classical video prediction methods in terms of their ability to forecast future scenes but also delve into its performance compared to the ``predict and track" sequential prediction paradigm. For the comparison, we initially train the video prediction network with other configurations unchanged. Subsequently, we employ the state-of-the-art trackers, SiamBAN\cite{9847396} and MixFormer\cite{cui2023mixformer}, to track the predicted scenes and locate the target in future scenes. We conduct experiments on the UAV123VP and VisDroneVP datasets, reporting results for two settings: predicting the future 8 frames based on 8 input frames (8→8) and predicting the future 8 frames based on 4 input frames (4→8).}
\subsubsection{Quantitative Analysis}

\begin{table*}[ht]
\centering
{
\renewcommand\arraystretch{1.15}
\caption{COMPARISONS WITH STATE-OF-THE-ART METHODS ON UAV123VP}
\resizebox{0.8\linewidth}{!}{
\begin{tabular}{l|c|ccccc|rc|c}
\Xhline{1pt}
\rowcolor[HTML]{EFEFEF} 
{Method} & {T→T'}         & {MSE$\downarrow$}     & {MAE$\downarrow$}      & {SSIM$\uparrow$}  & {PSNR$\uparrow$}  & {LPIPS$\downarrow$} & {Flops(G)} & {FPS}  & {Params(M)}\\ \Xhline{1pt}
ConvLSTM\cite{shi2015convolutional}        &                       & 1831.59          & 11658.29          & 0.499          & 21.84          & 0.719          & 760.83           & 28  &  15.50     \\
PredRNN\cite{wang2017predrnn}         &                       & 2067.84          & 12615.69          & 0.469          & 21.31          & 0.707          & 1545.22          & 14       &  24.56     \\
PredRNN++\cite{wang2018predrnn++}       &                       & 1894.31          & 11768.90           & 0.503          & 21.76          & 0.684          & 2267.14          & 10    &  39.31      \\
MIM\cite{wang2019memory}             &                       & 1794.42          & 11586.79          & 0.493          & 21.88          & {\ul 0.666}    & 2354.18          & 8      &  47.61\\
PhyDNet\cite{guen2020disentangling}         &                       & 2419.94          & 15006.74          & 0.448          & 20.09          & 0.836          & 198.66  & 36       &  3.09\\
PredRNNv2\cite{wang2022predrnn}       &                       & 1934.36          & 12249.48          & 0.474          & 21.50           & 0.697          & 1553.41          & 13   &  24.58  \\
SimVP\cite{gao2022simvp}           &                       & \textbf{1592.11} & {\ul 10564.57}    & {\ul 0.524}    & {\ul 22.47}    & 0.688          & 240.64            & 30      & 43.14\\
SimVPv2\cite{tan2022simvp}         &                       & 1673.00             & 10947.75          & 0.517          & 22.25          & 0.688          & 450.56            & 20   & 50.79   \\ 
\textbf{Ours}   & \multirow{-9}{*}{8→8} & {\ul 1618.44}    & \textbf{10456.99} & \textbf{0.535} & \textbf{22.54} & \textbf{0.641} &  399.36      & 19  &  94.22           \\ \hline
ConvLSTM\cite{shi2015convolutional}        &                       & 2004.64          & 12268.18          & 0.484          & 21.53          & 0.748          & 558.08            & 38 &  15.50     \\
PredRNN\cite{wang2017predrnn}         &                       & 1949.05          & {\ul 10902.78}          & 0.493          & 21.72          & 0.677          & 1133.57          & 18 &  24.56     \\
PredRNN++\cite{wang2018predrnn++}       &                       & 1997.45          & 12400.50           & 0.470           & 21.40           & 0.694          & 1661.95          & 13  &  39.31    \\
MIM\cite{wang2019memory}             &                       & 1726.02          & 11152.24          & 0.508          & 22.14          & \textbf{0.628}          & 1715.20            & 12 & 47.61      \\
PhyDNet\cite{guen2020disentangling}         &                       & 2596.78          & 15580.72          & 0.432          & 19.87          & 0.838          & 145.41           & 49  &  3.09     \\
PredRNNv2\cite{wang2022predrnn}       &                       & 1946.81          & 12241.26          & 0.479          & 21.59          & 0.663          & 1138.69          & 17       &   24.58    \\
SimVP\cite{gao2022simvp}           &                       & {\ul 1680.48}          & 11037.19          & {\ul 0.512}          & {\ul 22.23}          & 0.696          & 194.56            & 38 & 43.14      \\
SimVPv2\cite{tan2022simvp}         &                       & 1701.03          & 11098.14          & {\ul 0.512}          & 22.22          & 0.679          & 257.02           & 32    &   50.79      \\ 
\textbf{Ours}   & \multirow{-9}{*}{4→8}                   & \textbf{1631.82} &\textbf{10614.33} & \textbf{0.528}  &   \textbf{22.48} &  {\ul 0.643}      &  459.78         &   20     &   94.22        \\ 
\Xhline{1pt}
\end{tabular}
}
}
\label{tab:1}
\end{table*}

\begin{table}[ht]
\centering
\renewcommand\arraystretch{1.15}
{
\caption{The motion prediction performance of TAFormer and tacking performance of state-of-the-arts tackers on UAV123VP}
\resizebox{0.95\linewidth}{!}{
\begin{tabular}{c|c|ccc|c}
\hline
\rowcolor[HTML]{EFEFEF} 
Method                      & T→T' & ROI-MSE$\downarrow$ & mIoU$\uparrow$ & ADE$\downarrow$ & FPS$\uparrow$                \\ \hline
                            & 8→8  &   40.46   &   0.806   & 1.221    &      13              \\
\multirow{-2}{*}{VP+SiamBAN\cite{9847396}}   & 4→8  &  41.53  &   0.769   &   1.729  & 14 \\ \hline
                            & 8→8  &   40.19   &  0.832    &   1.185  &   6    \\
\multirow{-2}{*}{VP+MixFormer\cite{cui2023mixformer}} & 4→8  &  40.41   &  0.804    &  1.446   & 7 \\ \hline
                            & 8→8  &   38.04 &   0.931   &  0.319   & 19         \\
\multirow{-2}{*}{Ours}      & 4→8  &   39.97 &   0.844   &  0.720   & 20         \\ \hline 
\end{tabular}
}
}
\label{tab:1.1}
\end{table}

\quad \\
\noindent \textbf{UAV123VP} \quad TABLE I presents the video prediction performance of our method and other approaches on UAV123VP. It is worth noting that methods with excellent prediction performance on natural scene and synthetic datasets may not perform as well on aerial video datasets. A significant factor is that aerial video datasets often involve videos captured by moving drones, where scenes include fast-moving objects and small targets, unlike natural scene datasets that are typically captured by stationary cameras, featuring slower-motion videos. As a result, the metrics on UAV datasets are generally lower compared to natural scene datasets\cite{tan2023openstl}. The proposed TAFormer effectively addresses these challenges, surpassing other video prediction methods and establishing new state-of-the-art performance on the majority of metrics in both configurations. Specifically, on the 8→8 setting, our method achieved an SSIM score of 0.535 and a PSNR score of 22.54, surpassing all previous state-of-the-arts. As the number of historical frames is reduced, the performance of all methods tends to decline, while our method consistently maintains optimal performance.

{In addition, our method exhibits a moderate level of computational complexity and inference speed, but has a large number of parameters. There are two main reasons for this: firstly, TAFormer employs ViT as the backbone, which inherently consumes a significant number of parameters due to its architecture; secondly, our approach encompasses modeling both the video frames and the motion states of the targets. Overall, our model accomplishes unified modeling of scene and target motion states with only a marginal increase of parameters and computational complexity. This is also attributed to the efficient utilization of parameters achieved by sharing the multi-head attention mechanism between scene and target motion states modeling.}

{Moreover, the proposed TAFormer also excels at predicting the future motion states of the target without significantly increasing computational complexity. TABLE II presents the performance of different methods in the aspect of target motion prediction on UAV123VP. It's evident that the predictions generated by the ``predict and track" sequential target-aware video prediction paradigm are consistently weaker than our "parallel" prediction approach. The reason is that methods following the sequential prediction paradigm predict future scenes first and then track the target in the predicted frames, experience increased uncertainty in target appearance and position with the blurring of predicted frames. This uncertainty adversely affects the tracker's performance, leading to a significant reduction in the credibility of the prediction results due to the accumulation of errors. Moreover, our method, benefiting from the unified modeling of historical scenes and target motion states along with the parallel prediction paradigm, exhibits higher inference speed.}

\begin{table*}[ht]
\centering
\renewcommand\arraystretch{1.15}
\caption{COMPARISONS WITH STATE-OF-THE-ART METHODS ON VISDRONEVP}
\resizebox{0.6\linewidth}{!}{
\begin{tabular}{l|c|ccccc}
\Xhline{1pt}
\rowcolor[HTML]{EFEFEF} 
{Method} & {T→T'}        & {MSE$\downarrow$}                   & {{MAE$\downarrow$}} & {SSIM$\uparrow$} & {PSNR$\uparrow$} & {LPIPS$\downarrow$}  \\ \Xhline{1pt}
ConvLSTM\cite{shi2015convolutional}        & \multirow{9}{*}{8→8} & 3713.29                        & {17628.67}                          & 0.380          & 18.31         & 0.709                 \\
PredRNN\cite{wang2017predrnn}         &                      & 2791.25   & {14344.31}                          & 0.460          & 19.84         & 0.539                 \\
PredRNN++\cite{wang2018predrnn++}       &                      & 3082.78   & {15702.58}                          & 0.400           & 19.13         & 0.664                 \\
MIM\cite{wang2019memory}             &                      & {\ul 2082.30}    & {11596.55}                          & 0.565         & 21.63         & 0.362                  \\
PhyDNet\cite{guen2020disentangling}         &                      & 3354.43   & {17772.21}                          & 0.385         & 18.47         & 0.665                \\
PredRNNv2\cite{wang2022predrnn}       &                      & 2351.76   & {12909.17}                          & 0.494         & 20.58         & 0.485                 \\
SimVP\cite{gao2022simvp}           &                      & 2241.28   & {12031.35}                          & 0.575         & 21.43         & \textbf{0.353}                \\
SimVPv2\cite{tan2022simvp}         &                      & 2151.51   & {{\ul 11666.61}}                          & {\ul 0.580}         & {\ul 21.74}         &  0.380                \\
\textbf{Ours}   &                      & \textbf{2070.76} & {\textbf{11384.70}}                         & \textbf{0.592}     & \textbf{21.98}     & {\ul 0.364}         \\ \hline
ConvLSTM\cite{shi2015convolutional}        & \multirow{9}{*}{4→8} & 3397.42  & {16582.69}                          & 0.387         & 18.72         & 0.692                 \\
PredRNN\cite{wang2017predrnn}         &                      & 2819.43   & {14431.64}                          & 0.445         & 19.80          & 0.587                \\
PredRNN++\cite{wang2018predrnn++}       &                      & 2928.47   & {14965.92}                          & 0.417         & 19.42         & 0.636               \\
MIM\cite{wang2019memory}             &                      & 2402.86   & {13028.75}                          & 0.489         & 20.59         & 0.489                \\
PhyDNet\cite{guen2020disentangling}         &                      & 3763.34   & {18973.22}                          & 0.373         & 17.90        & 0.689                   \\
PredRNNv2\cite{wang2022predrnn}       &                      & 3926.89   & {17492.97}                          & 0.366         & 17.88         & 0.580                  \\
SimVP\cite{gao2022simvp}           &                      & 2291.84   & {12257.91}                          & 0.560          & 21.32         & 0.417                 \\
SimVPv2\cite{tan2022simvp}         &                      & {\ul 2213.89}   & {{\ul 12093.86}}                          & {\ul 0.568}         & {\ul 21.43}         & {\ul 0.396}                 \\
\textbf{Ours}   &                      & \textbf{2127.95} & {\textbf{11564.08}}                         & \textbf{0.589}     & \textbf{21.85}     & \textbf{0.365}      \\ \Xhline{1pt}
\end{tabular}
}
\label{tab:2}
\end{table*}

% \\ \hline
% \multirow{2}{*}{\textbf{Motion Metrics}} & 8→8           & \multicolumn{1}{c|}{\textbf{ROI-MSE}} & \multicolumn{1}{c|}{\textbf{30.09}} & \multicolumn{1}{l|}{\multirow{2}{*}{\textbf{mIoU}}} & \multicolumn{1}{c|}{\textbf{0.860}} & \multicolumn{1}{c|}{\multirow{2}{*}{\textbf{ADE}}} & \multicolumn{1}{c}{\textbf{0.479}} \\ \cline{2-2} \cline{4-4} \cline{6-6} \cline{8-8} 
%                                          & 4→8           & \multicolumn{1}{c|}{($\times 10^{-3}$)}                                         & \multicolumn{1}{c|}{\textbf{30.37}} & \multicolumn{1}{l|}{}                               & \multicolumn{1}{c|}{\textbf{0.472}} & \multicolumn{1}{c|}{}                              & \multicolumn{1}{c}{\textbf{2.311}} 

\begin{table}[ht]
\centering
\renewcommand\arraystretch{1.15}
{
\caption{The motion prediction performance of TAFormer and tacking performance of state-of-the-arts tackers on VisDroneVP}
\resizebox{0.95\linewidth}{!}{
\begin{tabular}{c|c|ccc|c}
\hline
\rowcolor[HTML]{EFEFEF} 
Method                      & T→T' & ROI-MSE$\downarrow$ & mIoU$\uparrow$ & ADE$\downarrow$ & FPS$\uparrow$                \\ \hline
                            & 8→8  &   30.16   &   0.743   & 1.168    &       13             \\
\multirow{-2}{*}{VP+SiamBAN\cite{9847396}}   & 4→8  &  30.43  &  0.429   &   2.923  & 14 \\ \hline
                            & 8→8  &   30.14   &  0.742    &   1.354  &    6   \\
\multirow{-2}{*}{VP+MixFormer\cite{cui2023mixformer}} & 4→8  &  30.29  &  0.431    &  2.605   & 7 \\ \hline
                            & 8→8  &   30.09 &   0.860   &  0.479   & 19         \\
\multirow{-2}{*}{Ours}      & 4→8  &   30.37 &   0.472   &  2.331   & 20         \\ \hline 
\end{tabular}
}
}
\label{tab:2.1}
\end{table}
\noindent \textbf{VisDroneVP} \quad TABLE \ref{tab:2} presents the performance of various methods on VisDroneVP under the configurations of 8→8 and 4→8. This dataset has a smaller volume (approximately $28.16\%$ of the training data in UAV123VP), allowing experiments to test the generalization of different methods on a small dataset. Additionally, compared to UAV123VP, this dataset contains more fast-moving objects, requiring the models to capture more complex temporal dynamics while maintaining the clarity of object appearances, posing a greater prediction challenge.

It is worth noting that methods like ConvLSTM\cite{shi2015convolutional} and PredRNN\cite{wang2017predrnn} show a significant performance gap on this dataset compared to UAV123VP. Besides not adapting well to a smaller dataset, part of the reason is that these methods lead to more loss of appearance information for fast-moving objects. Nevertheless, our method is not significantly affected by the small dataset due to the design of STA and ISM, aiding the model in fully exploring spatiotemporal dynamics in aerial videos and paying more attention to areas where the target locates. Moreover, TSGL helps to make the predictions around the region of interest more accurate. Specifically, our model achieves an SSIM of 0.592 and 0.589 for the 8→8 and 4→8 sets, respectively, surpassing previous state-of-the-arts. {TABLE IV presents the performance of different methods in target motion prediction on VisDroneVP. For the prediction of target states of TAFormer, the mIoU of the bounding boxes reaches 0.86 in the 8→8 configuration, indicating highly accurate predictions. However, in the 4→8 setting, when the historical context is reduced, there is a significant impact on predicting the target's motion state, resulting in an mIoU of only 0.472. This is largely due to the rapid movement of the targets in VisDroneVP, making it challenging to predict accurately with less context. Nevertheless, TAFormer's predictive performance for target motion states also surpasses that of the ``predict and track" sequential target-aware video prediction paradigm.}

\subsubsection{Qualitative Analysis}
\begin{figure*}[t]
	\setlength{\abovecaptionskip}{1pt}
	\centering
	\includegraphics[width=0.9\linewidth]{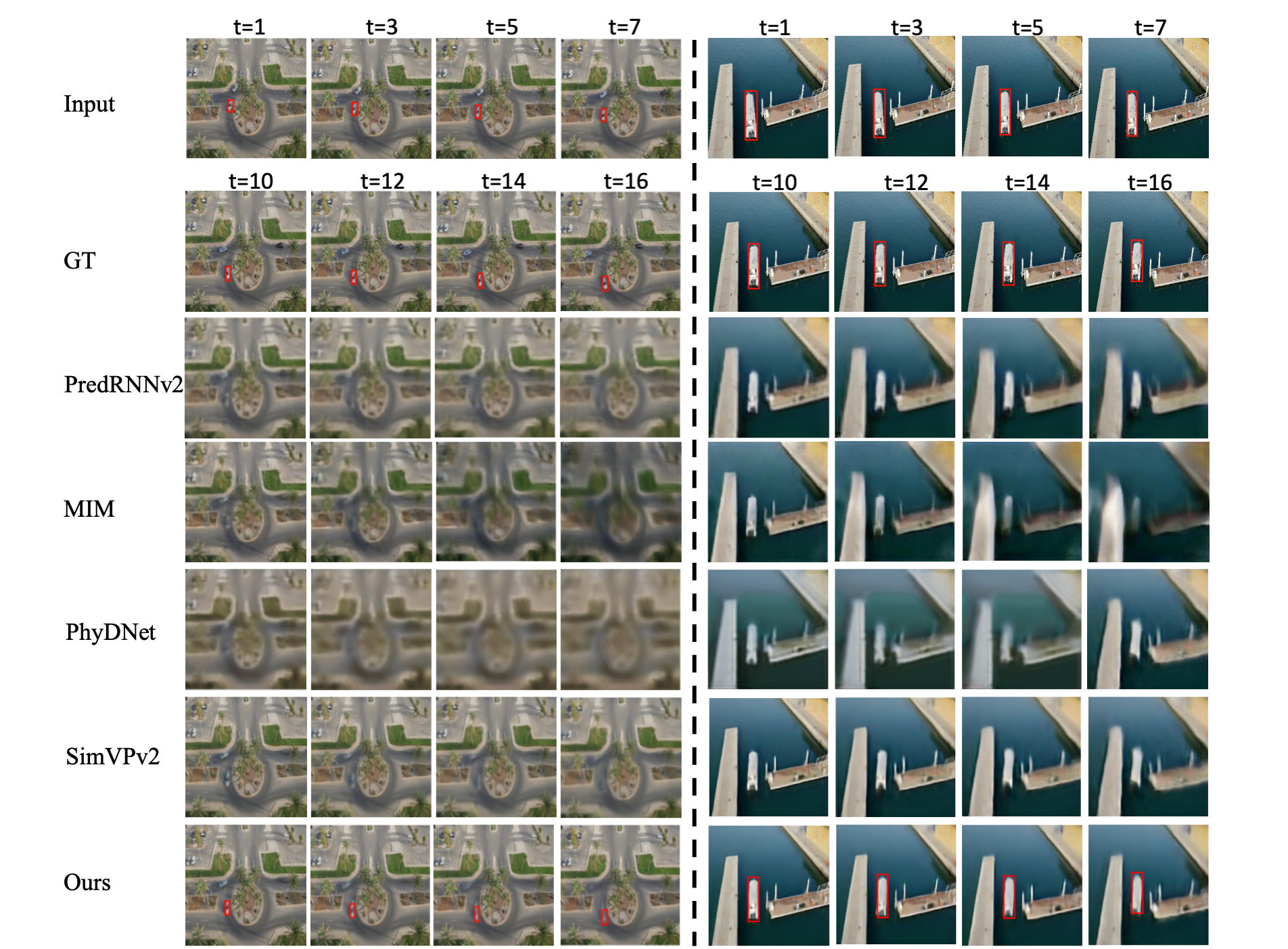}
	\caption{\textbf{Results of qualitative analysis for our TAFormer and other comparative methods in the 8→8 configuration.} The first 2 rows shows input frames and ground truths of two samples, while each set of four consecutive images below showcases the prediction results of a specific method. Our approach additionally highlights the location of the object of interest with red bounding boxes based on the prediction results.}
	\label{fig:compare}
\end{figure*}

To better analyze the prediction performance of different models, we visualize the prediction results for a typical small target with non-linear motion and a departing ship during the testing phase, as shown in Fig. \ref{fig:compare}. 

It is evident that methods that perform well in natural scene datasets encounter challenges when directly applied to aerial video datasets, which is manifested in increasingly blurry predictions over time and the problem of losing the region of interest due to the blurriness. RNN-based methods (PredRNNv2\cite{wang2022predrnn}, MIM\cite{wang2019memory} and PhyDNet\cite{guen2020disentangling}) show decent performance in predicting nearby video frames, but as time progresses, influenced by the limitations of the network architecture, prediction errors accumulate. The models gradually forget and lose memory information from earlier video frames, causing the background to become blurry and distorted. The region of interest gradually turns into a ghost, and by the last time step, even the target's contour is difficult to discern, especially for the left case in Fig. \ref{fig:compare}. In addition, the predictions of MIM\cite{wang2019memory} and PhyDNet\cite{guen2020disentangling} also suffer from brightness deviations in video frames. The recent state-of-the-art work, SimVPv2\cite{tan2022simvp}, abandons the recurrent neural network architecture and utilize a pure CNN architecture with an Encoder-Translator-Decoder structure. It achieves good prediction results through a non-autoregressive prediction approach, predicting multiple frames at once and provides relatively clear background predictions. The error accumulation problem of the RNN architecture is mitigated to some extent. However, due to the lack of special attention to the region of interest, the final frame's prediction still struggles to discern the region of interest. Overall, our method not only maintains the clarity of the background but also predicts the position of the region of interest. Moreover, the appearance and contour of the region of interest are preserved until the last time step.

\begin{figure}[t]
	\setlength{\abovecaptionskip}{1pt}
	\centering
	\includegraphics[width=1.0\linewidth]{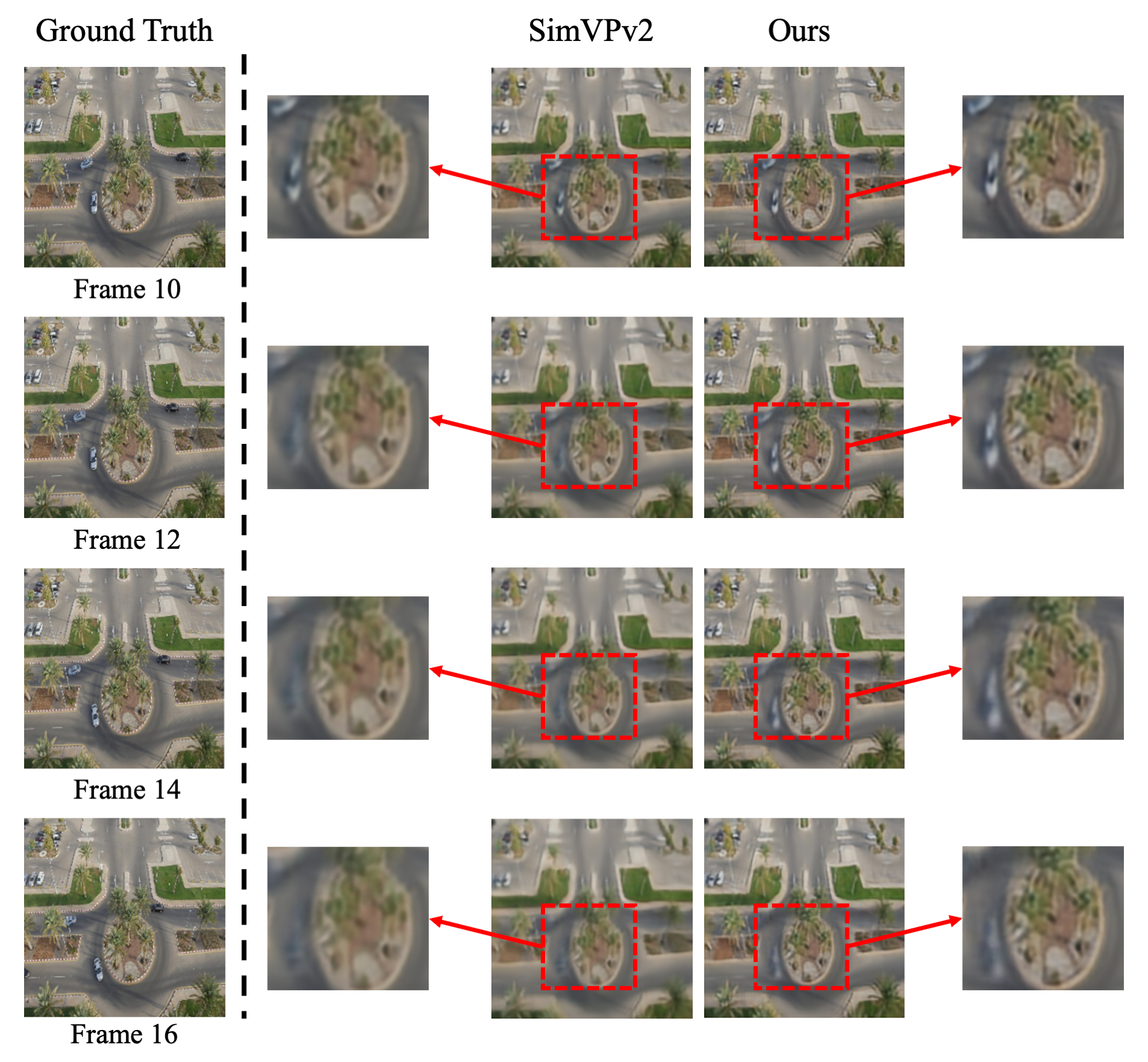}
	\caption{\textbf{Qualitative analysis of the prediction results of SimVPv2 and our method.} The contents within the red dashed boxes are enlarged to highlight the prediction effects around the target.}
	\label{fig:comvis}
\end{figure}

To provide a clearer and more intuitive illustration of the prediction sensitivity of our approach to the interested target and its motion, we present magnified images of the prediction results of SimVPv2\cite{tan2022simvp} and our method in Fig. \ref{fig:comvis}. Clearly, as time progresses, our TAFormer better preserves the predictions of the appearance of the target and its surrounding area. In the prediction at the 12th frame, the target is about to disappear in the prediction result of SimVPv2, while our result still discern the shape outline of the car. In the subsequent prediction results, SimVPv2 is unable to anticipate the outline and position of the target, whereas our method maintains the appearance information of the car relatively well. Even though predictions become blurry at the 14th and 16th frames, the car's movement trend around the roundabout is accurately predicted.

\subsection{Ablation Studies} \label{exp:abl}
In this section, ablation studies are conducted on various modules of the proposed TAFormer, the initialization methods of messengers, and the number of messengers for both types. Unless otherwise specified, all experiments are conducted on the UAV123VP dataset with an 8→8 configuration.
\subsubsection{Ablation Studies on Modules}
We conduct a series of ablation studies to explore the impact of each module on our model's prediction performance. The results are presented in TABLE \ref{tab:3}. {VP represents Video Prediction, which is the baseline model and utilizes the classical Transformer Encoder following \cite{vaswani2017attention} instead of the TAFormer Encoder and solely predicts future video frames.} MP denotes Motion Prediction, which refers to the module related to bounding box prediction. 
% \begin{table}[ht]
% \centering
% \renewcommand\arraystretch{1.25}
% \caption{Ablation studies on the Modules.}
% \resizebox{\linewidth}{!}{
% \begin{tabular}{cccc|cccc}
% \hline 
% \rowcolor[HTML]{EFEFEF} 
% MP & {STA} & {ISM} & {TSGL} & {MSE} & {SSIM} & {ROI-MSE} & {mIoU} \\ \hline
%             &              &              &               & 1712.71      & 0.517         & —               & —            \\
%             & $\checkmark$            &              &               & 1656.37      & 0.527         & —               & —            \\ \hline     
% $\checkmark$           & $\checkmark$            &              &               & 1640.47      & 0.534         & 38.24            & 0.892         \\
% $\checkmark$           & $\checkmark$            & $\checkmark$            &               & 1617.82      & 0.535         & 38.11            & 0.925         \\ \hline
% $\checkmark$           & $\checkmark$            & $\checkmark$            & $\checkmark$             & 1618.44      & 0.535         & 38.04            & 0.931   \\     \hline
% \end{tabular}
% }
% \label{tab:3}
% \end{table}

% Please add the following required packages to your document preamble:
% \usepackage[table,xcdraw]{xcolor}
% Beamer presentation requires \usepackage{colortbl} instead of \usepackage[table,xcdraw]{xcolor}

\begin{table}[ht]
\centering
\renewcommand\arraystretch{1.25}
{
\caption{Ablation studies on the Modules.}
\resizebox{\linewidth}{!}{
\begin{tabular}{ccccc|cccc}
\hline
\rowcolor[HTML]{EFEFEF} 
VP           & MP           & STA          & ISM                  & TSGL         & MSE     & SSIM  & ROI-MSE & mIoU  \\ \hline
$\checkmark$ &              &              &                      &              & 1712.71 & 0.517 & —       & —     \\
             & $\checkmark$ &              & \multicolumn{1}{l}{} &              & —       & —     & —       & 0.866 \\
$\checkmark$ &              & $\checkmark$ &                      &              & 1656.37 & 0.527 & —       & —     \\
$\checkmark$ & $\checkmark$ & $\checkmark$ &                      &              & 1640.47 & 0.534 & 38.24   & 0.892 \\
$\checkmark$ & $\checkmark$ & $\checkmark$ & $\checkmark$         &              & 1617.82 & 0.535 & 38.11   & 0.925 \\ \hline
$\checkmark$ & $\checkmark$ & $\checkmark$ & $\checkmark$         & $\checkmark$ & 1618.44 & 0.535 & 38.04   & 0.931 \\ \hline
\end{tabular}
}}
\label{tab:3}
\end{table}

The baseline model lacks the ability to predict target future motion states and can't adequately explore the spatiotemporal dynamics in videos. As a result, the prediction performance is poor, with an SSIM of only 0.517. MP introduces certain target motion information into the model. By predicting future motion states of the target from its historical trajectory, the model can better understand the dynamic behavior of the target, aiding in the more accurate capture of target motion details in video prediction. Compared to the baseline, the introduction of MP has led to some improvement in the model's scene prediction performance, with SSIM increasing to 0.530. While the model with only MP has gained the ability to predict the future motion states of the target, the motion prediction metric is relatively low, with mIoU reaching only 0.743. When STA is introduced alone, the spatial static attention and temporal dynamic attention allow the model to more comprehensively explore the spatiotemporal dynamics in videos and better model the appearance and motion of the scenes.  In comparison to the baseline, the introduction of STA results in an improvement in SSIM, reaching 0.527. Simultaneously introducing MP and STA will further enhance the model's ability to predict scene and target motion, with an increase in SSIM to 0.534 and mIoU to 0.892. On this basis, if we further introduce ISM, the model's scene prediction capability will experience a marginal improvement, with SSIM increasing to 0.535. Simultaneously, the model's target motion prediction ability will see a significant enhancement, with mIoU increasing to 0.925. ISM further strengthens the interaction between video and target motion information. By effectively transmitting information, the model can better consider the relationship between video evolution and target states. The introduction of TSGL makes the model pay more attention to the position and content of the target, further enhancing the model's ability to predict target motion states and resulting in a further increase in mIoU to 0.931.

\begin{figure}[ht]
	\setlength{\abovecaptionskip}{1pt}
	\centering
	\includegraphics[width=1.0\linewidth]{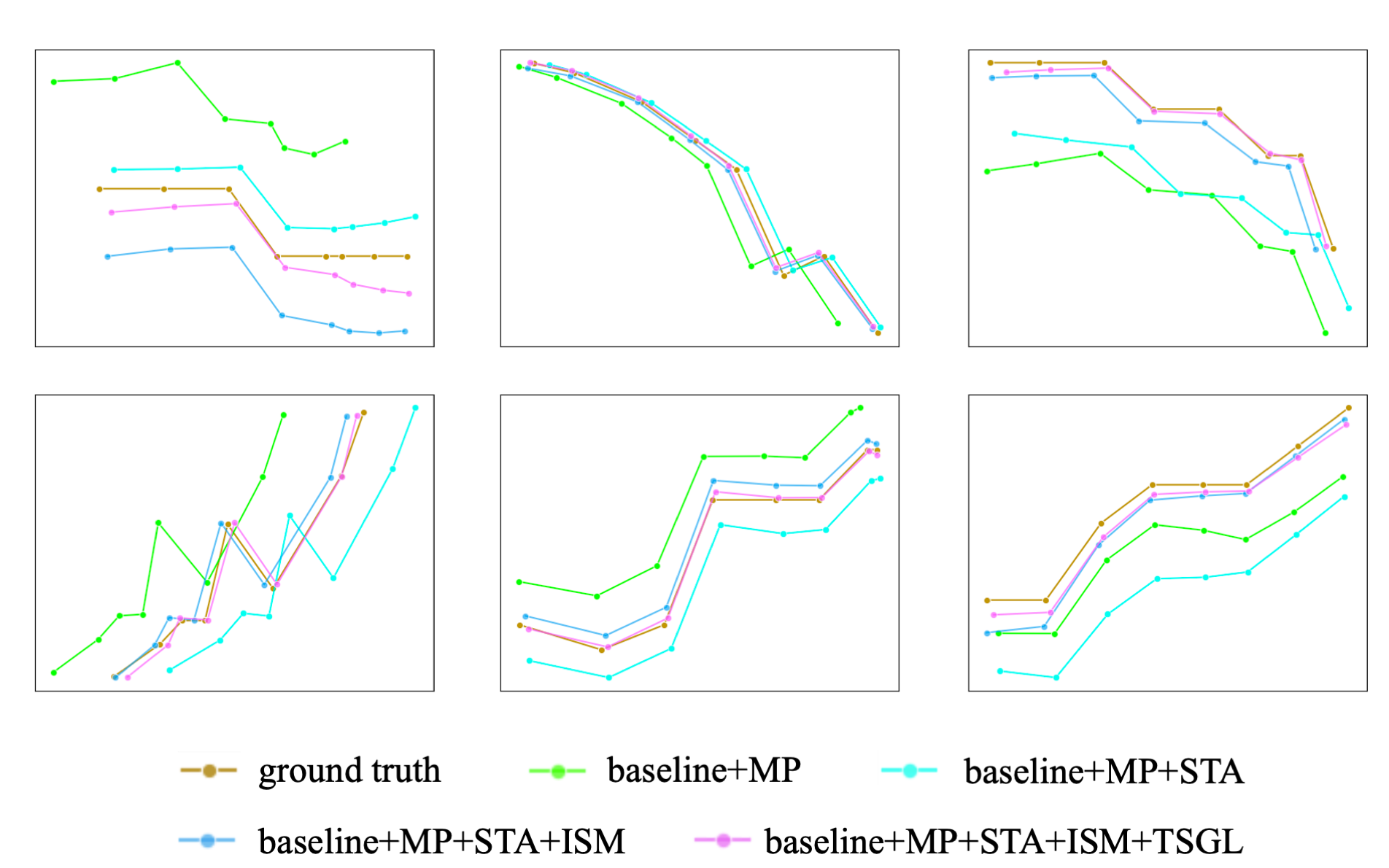}
	\caption{Examples of predicted trajectories of different settings on the UAV123VP dataset.}
	\label{fig:ablation}
\end{figure}

To visually demonstrate the role of each module in predicting the future states of the target, we select several samples and obtain predicted trajectories using models with different settings, as shown in Fig. \ref{fig:ablation}. Clearly, the addition of each module results in predicted trajectories that are closer to the ground truth.

\subsubsection{Ablation Studies on Messengers Initialization}
In this section, we investigate the impact of the initialization sources of two types of messengers on the model's prediction performance. We employ two methods to initialize $T_R$: random initialization and initialization using features from the region of interest. Similarly, we use two methods to initialize $T_S$: random initialization and initialization using features from the target's historical motion state. The specific initialization methods are described in Section \ref{subsubsec:init}. The performance of the model with different initialization combinations is shown in Table \ref{tab:3.5}, where $I(T_R)$ and $I(T_S)$ represent the initialization methods for $T_R$ and $T_S$, respectively. When both $T_R$ and $T_S$ are randomly initialized, the SSIM of the predicted video frames is 0.531. The performance of predicting the target bounding box is relatively poor, with mIoU reaching only 0.705. Individually using ROI to initialize $T_R$ and utilizing historical motion states to initialize $T_S$ both lead to significant performance improvement. This is because, during the information gathering stage, initializing messenger tokens with environmental and motion information enables them to efficiently collect relevant information during the information collection phase. When the initialization of both messengers no longer relies on random values but is instead performed using ROI and historical motion states, the model's predictive performance is further enhanced. The SSIM for predicted video frames reaches 0.535, and the mIoU for the predicted target motion states reaches 0.931.
% Please add the following required packages to your document preamble:
% \usepackage[table,xcdraw]{xcolor}
% Beamer presentation requires \usepackage{colortbl} instead of \usepackage[table,xcdraw]{xcolor}
\begin{table}[ht]
\centering
\renewcommand\arraystretch{1.15}
\caption{Ablation Studies on Sources of Initial Messengers.}
\resizebox{0.8\linewidth}{!}{
\begin{tabular}{cc|ccc}
\hline
\rowcolor[HTML]{EFEFEF} 
$I(T_R)$ & $I(T_S)$ & {MSE} & {SSIM} & {mIoU} \\ \hline
random      & random      & 1642.09      & 0.531         & 0.705         \\
ROI         & random      & 1640.30       & 0.532         & 0.908         \\
random      & states      & 1646.45      & 0.532         & 0.901         \\ \hline
ROI         & states      & 1618.44      & 0.535         & 0.931         \\ \hline
\end{tabular}
}
\label{tab:3.5}
\end{table}

\subsubsection{Ablation Studies on Number of Messengers}
In target-aware video prediction, the information processed in video frame prediction (predicting the pixel values of each future frame) is far greater than that in target motion prediction (predicting only the sequence of bounding boxes). In the design of ISM, the quantity and ratio of the two messengers will directly affect the efficiency of information collecting and passing. In this section, we conduct experiments on the quantities of $T_R$ and $T_S$, and the results are shown in TABLE \ref{tab:4}.
\begin{table}[ht]
\centering
\renewcommand\arraystretch{1.15}
\caption{Ablation Studies on Number of Messengers}
\begin{tabular}{ccc|cccc}
\hline
\rowcolor[HTML]{EFEFEF} 
\textbf{$\#T_R$} & \textbf{$\#T_S$} & \textbf{$\frac{\#T_R}{\#T_S}$} & \textbf{MSE}     & \textbf{SSIM}  & \textbf{ROI-MSE} & \textbf{mIoU}  \\ \hline
2           & 2           & 1              & 1638.33  &  0.532 &  38.25  &   0.927        \\
4           & 4           & 1              & \textbf{1616.06}          & 0.534          & 38.27            & 0.733          \\
8           & 8           & 1              & 1632.60          & 0.534 & 38.33 &  0.704 \\ \hline
4           & 2           & 2              & 1629.36          & 0.531          & 38.24            & 0.897          \\
8           & 4           & 2              & 1630.92          & 0.534          & 38.29            & 0.760           \\ 
16          & 8           & 2              & 1647.22          & 0.533          & 38.28 &0.926       \\ \hline
4           & 1           & 4              & 1623.07          & 0.533  & 38.18  & 0.911 \\
8           & 2           & 4              & 1618.44          & \textbf{0.535} & \textbf{38.04}  & \textbf{0.931} \\
16          & 4           & 4              & 1645.58          & 0.533          & 38.25            & 0.930           \\  \hline
\end{tabular}%
\label{tab:4}
\end{table}

% \begin{table}[ht]
% \centering
% \renewcommand\arraystretch{1.15}
% \caption{}
% \begin{tabular}{ccc|cccc}
% \hline
% \rowcolor[HTML]{EFEFEF} 
% \textbf{$\#T_R$} & \textbf{$\#T_S$} & \textbf{$\frac{\#T_R}{\#T_S}$} & \textbf{MSE}     & \textbf{SSIM}  & \textbf{ROI-MSE} & \textbf{mIoU}  \\ \hline
% 4           & 4           & 1              & \textbf{1616.06}          & 0.534          & 38.27            & 0.733          \\

% 4           & 2           & 2              & 1629.36          & 0.531          & 38.24            & 0.897          \\
% 8           & 4           & 2              & 1630.92          & 0.534          & 38.29            & 0.760           \\ 
% 8           & 2           & 4              & 1618.44          & \textbf{0.535} & \textbf{38.04}  & \textbf{0.931} \\
% 16          & 4           & 4              & 1645.58          & 0.533          & 38.25            & 0.930    
%            \\  \hline
% \end{tabular}%
% \label{tab:4}
% \end{table}

It is easy to observe that the impact of the number and ratio of messengers on video frame prediction is relatively weak, mainly manifested in slight fluctuations in the MSE and SSIM metrics. And when the quantity of $T_R$ is set to 8, regardless of the quantity set for $T_S$, the prediction accuracy of video frames can achieve relatively satisfactory results. However, for the prediction of target motion states, the ratio between the two messengers has a significant influence. When the ratio $\frac{\#T_R}{\#T_S}$ is large, the mIoU metric consistently stays high; however, when this ratio is small, the mIoU metric becomes unstable and relatively lower. This is primarily because when the quantity of $ T_R $ is not overwhelmingly superior to $ T_S $, as mentioned earlier, there is a mismatch in the information between predicting video frames and target boxes. This leads to inefficiency in information transfer, preventing the model from effectively leveraging environmental information to predict the future states of the target. Overall, when the quantity of $T_R$ is set to 8 and $T_S$ is set to 2, the model achieves its best performance in video frame prediction and target motion state prediction, with an SSIM of 0.535 and an mIoU of 0.931.

\subsubsection{{Ablation Studies on the three parts of ISM}}
{
The introduction of ISM aims to facilitate the interaction and mutual learning between scene information and target motion state information. To verify the role of each component of ISM, we conducted ablation experiments on the three parts of it, and the results are shown in the Table \ref{tab:m1}, where (a), (b), and (c) respectively represent the Messengers Initialization, Message Collecting, and Message Passing phases, with a checkmark indicating the utilization of that particular part. Configurations marked with (b) and (c) indicate that the initialization of the messengers in phase (a) used random values.}
    
\begin{table}[htbp]
\centering
{
\caption{Ablation studies on the three parts of ISM.}
\begin{tabular}{ccc|cccc}
\hline
\rowcolor[HTML]{EFEFEF} 
\textbf{(a)}          & \textbf{(b)}         & \textbf{(c)}          & \textbf{MSE}         & \textbf{SSIM}        & \textbf{ROI-MSE}     & \textbf{mIoU}        \\ \hline
  &  &  & 1650.48 & 0.532 & 38.10 & 0.889 \\
                     & $\checkmark$                    & $\checkmark$                     &     1642.09                 &       0.531              &      38.05                &         0.705             \\
$\checkmark$                     &                      & $\checkmark$                     &    1622.23       &   0.532        &       38.09               &        0.910              \\
$\checkmark$                     & $\checkmark$                   &                      &     1631.63        &          0.530            &        38.12         &       0.913               \\
$\checkmark$                     & $\checkmark$                    & $\checkmark$                     &     1618.44                 &       0.535               &      38.04                &         0.931             \\ \hline
\end{tabular}
\label{tab:m1}
}
\end{table}

{
It's evident that the absence of each part will result in performance loss. The variation in mIoU values reflects the significant impact of each part of ISM on target state prediction, especially when messengers are initialized randomly, resulting in an mIoU of only 0.705. When all three parts are present simultaneously, the model achieves peak performance in predicting both scene and target motion states.}

\subsection{Anaysis on Typical Cases} \label{exp:ana}
\begin{figure}[t]
	\setlength{\abovecaptionskip}{1pt}
	\centering
	\includegraphics[width=1.0\linewidth]{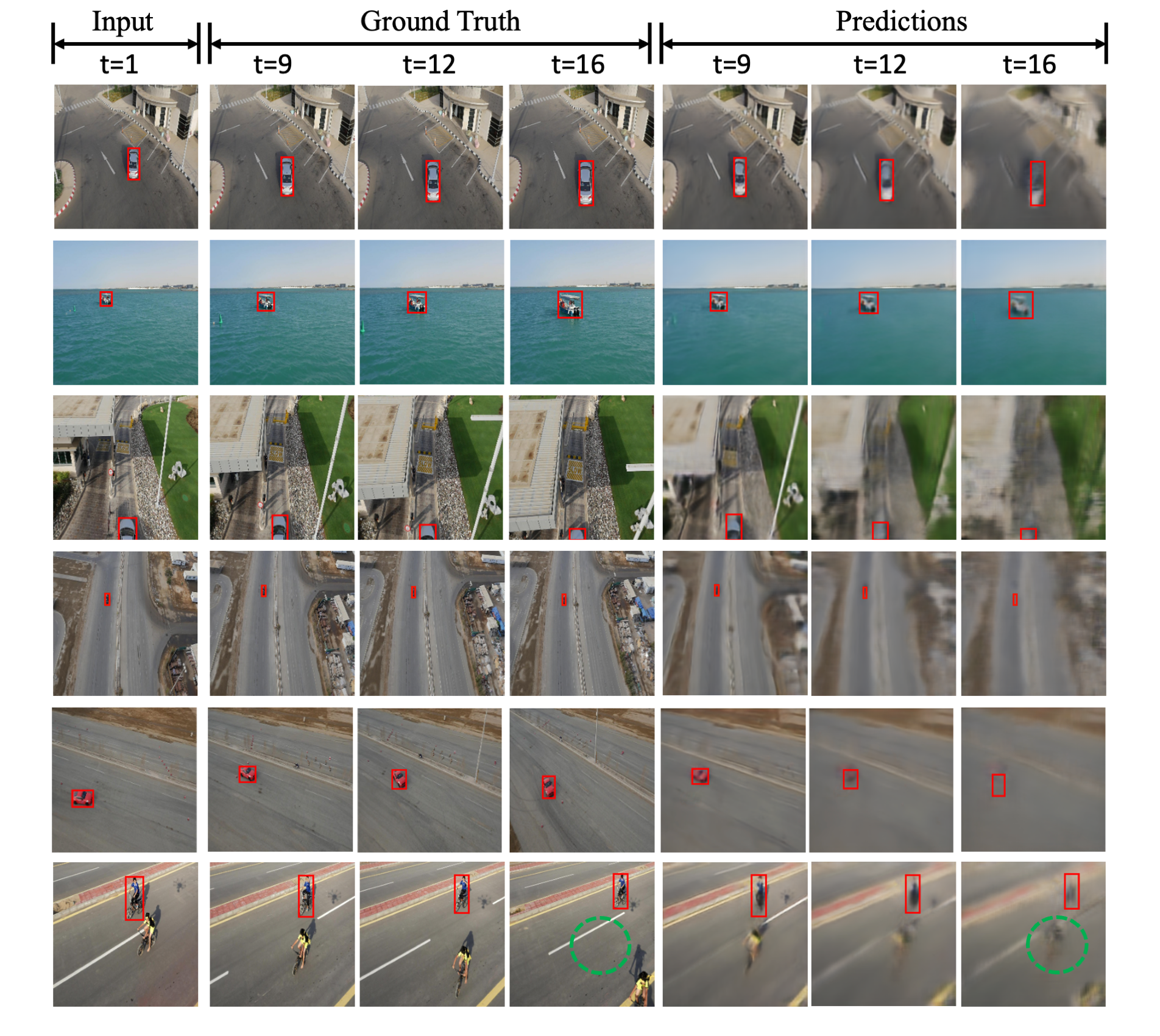}
	\caption{\textbf{Visualization of some special cases.} Each row represents a case, with the first column showing the input initial frame, columns 2-4 representing the ground truth, and columns 5-7 showing the predicted values by our TAFormer for the corresponding frames. }
	\label{fig:difficult}
\end{figure}

To gain a more comprehensive understanding of our model's prediction capabilities across various scenarios and special cases, we select representative scenes for visualization in Fig. \ref{fig:difficult}. {In the first two rows, the perspective effect is evident as the targets undergo changes in size and distance from the drone's position, illustrating a transition from far to near. In the third row, the region of interest is situated at the image's edge. In the fourth and fifth rows, there are significant background changes (note the position of the intersection in the first and last frames of the fourth row, and the position of the utility pole in the fifth row). Finally, in the fifth row, the target in the video moves rapidly, leading to substantial changes in their positions within the frames.}

Overall, our model performs accurate predictions of target positions, even in scenarios where the target is about to exit the frame (row 3) or when the perspective effect is prominent (row 2). {However, there are still some challenges in predicting video frames. Specifically, the phenomenon of background blurring appears early and is more pronounced when there are significant changes in the background between historical and future frames. This is largely attributed to the interference of modeling video motion information, affecting the preservation of both target and background appearance. In the last row, despite the target (wearing yellow clothes and riding a bicycle) having moved to the bottom right corner in the ground truth, the artifacts still persist in the circled region of the predicted frames. In future research, we aim to address these issues by exploring techniques to reduce the interference between learning video motion information and maintaining objects' appearance.}

\section{Conclusion}
In this paper, we propose a novel approach named TAFormer for target-aware aerial video prediction. TAFormer aims to predict the evolution of the environment and future states of the interested target conditioned on historical aerial video frames and the target's motion states. It goes beyond traditional research that rely solely on perception and tracking for aerial videos interpretation, overcoming limitations in meeting the demands for higher-level intelligent analysis of aerial videos. To validate the predictive performance of TAFormer in aerial scenarios, we have transformed two drone-captured SOT datasets, UAV123 and VisDrone, into target-aware aerial video prediction datasets, called UAV123VP and VisDroneVP. Through extensive experiments on the two datasets, TAFormer demonstrates outstanding performance, surpassing previous methods not only in achieving state-of-the-art performance in video prediction but also in adapting to the additional requirements of target awareness for aerial video interpretation. {However, there are limitations that need to be addressed in future work. Firstly, the model only supports the prediction of future motion states for a single target, thus lacking the capability to handle scenes with multiple targets of interest. Secondly, our approach assumes knowledge of the historical motion states of the target, which may not always be available in real-world scenarios where perception algorithms are required to locate targets in scenes. We will persist in refining our approach in future research to address failure cases and model limitations.}
% \begin{algorithm}[H]
% \caption{Weighted Tanimoto ELM.}\label{alg:alg1}
% \begin{algorithmic}
% \STATE 
% \STATE {\textsc{TRAIN}}$(\mathbf{X} \mathbf{T})$
% \STATE \hspace{0.5cm}$ \textbf{select randomly } W \subset \mathbf{X}  $
% \STATE \hspace{0.5cm}$ N_\mathbf{t} \gets | \{ i : \mathbf{t}_i = \mathbf{t} \} | $ \textbf{ for } $ \mathbf{t}= -1,+1 $
% \STATE \hspace{0.5cm}$ B_i \gets \sqrt{ \textsc{max}(N_{-1},N_{+1}) / N_{\mathbf{t}_i} } $ \textbf{ for } $ i = 1,...,N $
% \STATE \hspace{0.5cm}$ \hat{\mathbf{H}} \gets  B \cdot (\mathbf{X}^T\textbf{W})/( \mathbb{1}\mathbf{X} + \mathbb{1}\textbf{W} - \mathbf{X}^T\textbf{W} ) $
% \STATE \hspace{0.5cm}$ \beta \gets \left ( I/C + \hat{\mathbf{H}}^T\hat{\mathbf{H}} \right )^{-1}(\hat{\mathbf{H}}^T B\cdot \mathbf{T})  $
% \STATE \hspace{0.5cm}\textbf{return}  $\textbf{W},  \beta $
% \STATE 
% \STATE {\textsc{PREDICT}}$(\mathbf{X} )$
% \STATE \hspace{0.5cm}$ \mathbf{H} \gets  (\mathbf{X}^T\textbf{W} )/( \mathbb{1}\mathbf{X}  + \mathbb{1}\textbf{W}- \mathbf{X}^T\textbf{W}  ) $
% \STATE \hspace{0.5cm}\textbf{return}  $\textsc{sign}( \mathbf{H} \beta )$
% \end{algorithmic}
% \label{alg1}
% \end{algorithm}

\section*{Acknowledgment}
This work was supported by the National Natural Science Foundation of China under Grant 62331027, 62171436 and the Key Deployment Program of the Chinese Academy of Sciences: KGFZD-145-23-18.

\ifCLASSOPTIONcaptionsoff
\newpage
\fi

\bibliographystyle{ieeetr} 
\bibliography{main}

% \section{Biography Section}
% If you have an EPS/PDF photo (graphicx package needed), extra braces are
%  needed around the contents of the optional argument to biography to prevent
%  the LaTeX parser from getting confused when it sees the complicated
%  $\backslash${\tt{includegraphics}} command within an optional argument. (You can create
%  your own custom macro containing the $\backslash${\tt{includegraphics}} command to make things
%  simpler here.)
 
% \vspace{11pt}

% \bf{If you include a photo:}\vspace{-33pt}
% \begin{IEEEbiography}[{\includegraphics[width=1in,height=1.25in,clip,keepaspectratio]{fig1}}]{Michael Shell}
% Use $\backslash${\tt{begin\{IEEEbiography\}}} and then for the 1st argument use $\backslash${\tt{includegraphics}} to declare and link the author photo.
% Use the author name as the 3rd argument followed by the biography text.
% \end{IEEEbiography}

% \vspace{11pt}

% \bf{If you will not include a photo:}\vspace{-33pt}
% \begin{IEEEbiographynophoto}{John Doe}
% Use $\backslash${\tt{begin\{IEEEbiographynophoto\}}} and the author name as the argument followed by the biography text.
% \end{IEEEbiographynophoto}

\vfill

\end{document}